%% file: main.tex
\PassOptionsToPackage{verbose=silent}{microtype}   %
\documentclass[11pt, logo, onecolumn, copyright, colorlinks=true, allcolors=blue]{nvidiatechreport}

\usepackage{iftex} 
\ifXeTeX
  \microtypesetup{tracking=false}
\fi
\ifPDFTeX
  \usepackage[utf8]{inputenc}
  \usepackage[T1]{fontenc}
  \usepackage{textcomp}
  \newcommand{\dn}[1]{\textcolor{red}{[Devanagari text: compile with LuaLaTeX]}}
\else
  \usepackage{fontspec}
  \ifLuaTeX
    \newfontfamily\devanagarifont[Script=Devanagari,Path=fonts/,BoldFont=NotoSansDevanagari-Bold.ttf,Renderer=HarfBuzz]{NotoSansDevanagari-Regular.ttf}
  \else
    \newfontfamily\devanagarifont[Script=Devanagari,Path=fonts/,BoldFont=NotoSansDevanagari-Bold.ttf]{NotoSansDevanagari-Regular.ttf}
  \fi
  \newcommand{\dn}[1]{{\devanagarifont #1}}   %
\fi

\usepackage{xurl}
\usepackage{graphicx}
\usepackage{booktabs}
\usepackage{tabularx}
\usepackage{array}
\usepackage{makecell}
\usepackage{adjustbox}
\usepackage{multirow}
\usepackage{caption}
\usepackage{subcaption}
\usepackage[normalem]{ulem}
\usepackage[section]{placeins}
\usepackage{rotating}      %
\usepackage{pdflscape}     %
\usepackage{xltabular}     %
\usepackage{newunicodechar}
\usepackage{amsmath}
\usepackage{xltabular,booktabs}
\usepackage{pifont}
\usepackage{xcolor}
\usepackage{tikz}
\usepackage{upquote}
\usepackage[numbers,square]{natbib}

\usepackage{pifont}
\newcommand{\cmark}{\ding{51}}
\newcommand{\xmark}{\ding{55}}
\newcommand{\pmark}{\ding{109}}

\newcolumntype{C}[1]{>{\centering\arraybackslash}p{#1}}

\usepackage{listings}
\lstdefinestyle{prompt}{
  basicstyle=\small\ttfamily, columns=fullflexible, keepspaces=true,
  breaklines=true, breakatwhitespace=true, breakindent=1.5em,
  frame=single, framerule=0.3pt, rulecolor=\color{black!40},
  xleftmargin=2pt, xrightmargin=2pt, aboveskip=6pt, belowskip=10pt,
  showstringspaces=false, upquote=true,
  literate={—}{{---}}1 {×}{{$\times$}}1 {→}{{$\rightarrow$}}1}

\newunicodechar{→}{\ensuremath{\rightarrow}}
\newunicodechar{↑}{\ensuremath{\uparrow}}
\newunicodechar{↓}{\ensuremath{\downarrow}}
\newunicodechar{≥}{\ensuremath{\geq}}
\newunicodechar{−}{\ensuremath{-}}
\newunicodechar{∑}{\ensuremath{\sum}}
\newunicodechar{∈}{\ensuremath{\in}}
\newunicodechar{ℓ}{\ensuremath{\ell}}
\newunicodechar{Δ}{\ensuremath{\Delta}}
\newunicodechar{κ}{\ensuremath{\kappa}}
\newunicodechar{τ}{\ensuremath{\tau}}
\newunicodechar{²}{\textsuperscript{2}}
\newunicodechar{₂}{\textsubscript{2}}
\newunicodechar{°}{\textdegree}
\newunicodechar{…}{\dots}

\ifLuaTeX
  \usepackage{lua-ul}
\else
\fi

\newcommand{\subhead}[1]{\par\medskip\noindent\textbf{#1}\par\nobreak\smallskip}   %


\numberwithin{table}{section}
\numberwithin{figure}{section}
\makeatletter
\newcommand{\appendixtitles}{%
  \titleformat{\section}{\large\bfseries\headingfont}{Appendix \thesection.}{0.5em}{##1}[]}
\g@addto@macro\appendix{\appendixtitles}
\makeatother

\makeatletter
\renewcommand{\abscontent}{%
  \noindent{\headingfont Abstract}\par\vskip6pt
  \noindent\parbox{\dimexpr\linewidth}{\absfont \theabstract}%
  \@ifundefined{@keywords}{}{\vskip1em \noindent \keywordsfont Keywords: \@keywords}%
}
\makeatother

\newcommand{\NPIv}{\mbox{NPI-v1}}
\newcommand{\NPIN}{\mbox{NPI-N}}
\newcommand{\NPIR}{\mbox{NPI-R}}

\title{Scaling Synthetic Pre-training Data and Evaluation for Indic LLMs}  %
\author{\large NVIDIA and AI4Bharat}                          %
\date{}

\begin{document}
\begin{abstract}
Building capable language models for India requires scalable multilingual data and robust evaluation. We introduce Nemotron-Pretraining-Indic-v1, a synthetic pretraining corpus of approximately 2.83 trillion tokens covering all 22 scheduled Indian languages in native script and twelve in romanized form. It is the largest openly released Indic pretraining corpus. Using our GPU-accelerated pipeline, we translate curated English documents, assess translation quality, and repair low-quality outputs while preserving technical content and document structure and maintaining source traceability. Through continued-pretraining experiments across twelve languages, we show that combining our synthetic corpus with natural Indic text improves knowledge and reasoning over a public-corpus baseline. Additional experiments show how data volume, script coverage, and Indic training share affect English performance. We also introduce the Indic-LLM-Eval-Suite, a multilingual benchmark suite for evaluating Indic language models across chat, instruction following, reasoning, knowledge, code, and translation. The suite combines existing Indic benchmarks with task-aware translations of English benchmarks, using automated quality assessment, targeted correction, and native-speaker review in a human-in-the-loop pipeline. We use the suite to comprehensively benchmark contemporary open-weight Indic and multilingual models across capabilities and languages, revealing substantial variation across tasks and languages. Together, these resources provide a scalable foundation for developing and evaluating more capable Indic language models.
\end{abstract}
\maketitle

\input{paper/01_introduction}
\input{paper/02_related_work}

\input{paper/03_scope}

\input{paper/04_pipeline}
\input{paper/06_experiments}   %
\input{paper/05_evaluation}   %
\input{paper/12_conclusion}

\input{paper/appendix_A_prompts}
\input{paper/appendix_corpus}
\input{paper/appendix_C_translator_selection}
\input{paper/appendix_execution}
\input{paper/appendix_analysis}
\input{paper/appendix_contributors}

\bibliography{references}\bibliographystyle{references}  %
\end{document}

%% file: paper/01_introduction.tex
\section{Introduction}\label{sec:1}

In recent years, artificial intelligence has witnessed transformative progress, largely driven by advancements in large language models (LLMs) for English and other high-resource languages. Central to this rapid evolution is pre-training on vast, high-quality, and knowledge-rich corpora. This foundational pre-training infuses models with the extensive world knowledge necessary across every stage of the training pipeline—including instruction following, multi-step reasoning, human preference alignment, and safety alignment. Equally critical to this trajectory has been the availability of standardized, well-curated evaluation benchmarks and dynamic human arenas (e.g., Artificial Analysis \cite{artificialanalysis}, LMSYS Chatbot Arena \cite{chiang2024chatbot}) that systematically measure diverse capabilities and track field-wide progress.

However, non-English languages—particularly Indian languages—lag significantly behind in both foundational data and evaluation resources, severely constraining the development of performant native LLMs. For instance, while state-of-the-art open pre-training datasets like FineWeb contain over 14 trillion English words, the multilingual FineWeb2 corpus yields only 11.17 billion words for Hindi - highlighting a vast order-of-magnitude imbalance. Similarly, evaluation collections spanning multiple capabilities remain limited for regional context. While recent work has begun to address these gaps, achieving capability parity with English LLMs demands a step-function advance in curating high-quality pre-training data alongside comprehensive, multi-capability evaluation suites.

In this paper, we describe our effort to make rapid and significant progress towards supporting a linguistically and culturally diverse country like India, with a focus on covering the 22 constitutionally recognized languages spanning 4 distinct language families. To navigate this challenging landscape at scale, we adopt two core principles:

\begin{itemize}
    \item \textbf{Openness and Reproducibility:} Making foundational resources, benchmarks, and creation recipes open source so the broader research community can build and innovate upon them—preventing multiple entities from redundantly recreating basic resources and accelerating field-wide progress.
    \item \textbf{Translation as Force Multiplier:} Leveraging modern LLM-driven machine translation as a scaling mechanism to extend high-quality English data across multiple target languages efficiently. To ensure reliability, we pair this approach with an \textbf{automated pipeline for multi-stage translation quality checks} combined with targeted human-in-the-loop verification.
\end{itemize}

For many Indic languages, high-quality naturally occurring text remains scarce relative to the amount of knowledge-rich material available in higher-resource languages \cite{kreutzer2022quality, villalobos2024rundata}. Translation and synthetic generation provide practical mechanisms for expanding this coverage by transferring information from curated source material into target languages \cite{doshi2024translationese, joshi2025adapting-hindi-nemotron}. However, scaling such generation does not by itself guarantee useful pretraining data. Generated text can alter meaning \cite{guerreiro2023hallucinations, freitag2021mqm}, damage document structure, introduce repetitive or unnatural language, and, in the case of romanization, produce inconsistent script and spelling choices \cite{madhani2023aksharantar, madhani2023bhashaabhijnaanam}. A fluent output may therefore still be unsuitable for pretraining if it systematically loses information or departs substantially from natural target-language distributions \cite{artetxe2020artifacts, doshi2024translationese}. Constructing synthetic corpora at scale thus requires mechanisms for preserving provenance \cite{gebru2021datasheets, longpre2024provenance}, detecting failures, selectively repairing low-quality outputs \cite{lu2024mqmape}, and ultimately determining whether the retained data improves models when used for training \cite{li2024dclm, penedo2024fineweb}.

Evaluation presents a parallel challenge. Existing Indic benchmarks vary substantially in their language coverage, capabilities, task construction, and validation procedures \cite{doddapaneni2023indiccorp, singh2024indicgenbench, verma2025milu, kamath2025benchmarkinghindillms, jayakumar2026indicifeval}. Extending prominent English benchmarks through translation can broaden evaluation coverage, but translation errors may alter the task itself \cite{artetxe2020artifacts, zeng2025marcobench}. Moreover, linguistic correctness alone does not guarantee culturally grounded evaluation, since translated questions can retain assumptions specific to their source context \cite{singh2025globalmmlu}. Reliable measurement therefore requires attention not only to benchmark breadth, but also to task preservation, correction, human validation, and reproducible scoring.

We bring together large-scale pretraining data construction and comprehensive, reproducible evaluation to advance Indic LLMs. We construct \textbf{Nemotron-Pretraining-Indic-v1 (\NPIv{})}, a source-traceable Indic pretraining corpus spanning 22 Indian languages and more than \textbf{2.8T} high-quality tokens across native and romanized scripts. We additionally introduce the \textit{Indic-LLM-Eval-Suite}, an ongoing and extensible benchmark creation effort aimed at making Indic LLM evaluation more comprehensive, reliable, and reproducible across languages and capabilities. Together, these resources enable us to study the full path from multilingual data construction and quality control to continued pretraining and reliable measurement of the resulting model capabilities.

Our main contributions are as follows:

\begin{itemize}

    \item \textbf{Nemotron-Pretraining-Indic-v1 (\NPIv{}):} We construct a source-traceable Indic pretraining corpus containing approximately \textbf{2.8 trillion tokens} across 22 Indian languages, with broad coverage across languages, scripts, and domains.

    \item \textbf{Data Creation Pipeline:} We develop a reusable, large-scale synthetic data creation pipeline for translation, quality assessment, targeted repair, and romanization. We systematically analyze its behavior, documenting common errors, unsuccessful approaches, and observed failure modes to provide practical insights for large-scale multilingual data construction.

    \item \textbf{Data Utility through Continued Pretraining:} We conduct controlled continued-pretraining experiments to determine whether the constructed data produces downstream model improvements. We compare multiple corpus compositions under uniform training settings and quantify gains across Indic languages while measuring the retention of English and non-Indic capabilities.

    \item \textbf{Indic-LLM-Eval-Suite:} We introduce a broad and extensible evaluation suite for Indic LLMs spanning multiple languages and capabilities. The suite combines existing Indic benchmarks with carefully translated prominent English benchmarks. We develop benchmark-specific translation, evaluation, and correction procedures followed by native-speaker validation, and integrate the resulting suite into NeMo-Gym\footnote{\url{https://github.com/NVIDIA-NeMo/gym}} to enable standardized and reproducible evaluation across models.

\end{itemize}

We open-source the corpus, data-generation recipes, evaluation resources, and associated documentation to support further research and development in Indic language modeling. The resulting checkpoints are used primarily as experimental instruments for measuring data utility rather than as state-of-the-art model releases. Together, these resources establish a reproducible framework connecting Indic data construction, quality control, model training, and reliable capability evaluation.

%% file: paper/02_related_work.tex
\section{Related Work and Positioning}\label{sec:2}
\paragraph{Indic and multilingual language resources}\label{sec:2.1}

Open Indic resources have progressively expanded the foundations for language modeling. IndicNLPSuite introduced IndicCorp, IndicBERT, and IndicGLUE, while IndicCorp v2 subsequently expanded corpus coverage alongside IndicBERT v2 and IndicXTREME \cite{doddapaneni2023indiccorp}. IndicLLMSuite extended this resource-building effort to large language models through Sangraha, which combines curated and filtered natural text with translated and romanized material, supported by reusable pipelines for data cleaning and document-preserving translation \cite{khan2024indicllmsuite}. More recent efforts have further explored large-scale synthetic or mixed Indic pretraining data, while several model releases report substantial Indic training corpora that are not publicly available.

Table~\ref{tab:3.3} compares these resources along dimensions that are important for interpreting corpus scale and accessibility. Since different works report token counts using different tokenizers, the reported sizes should be treated as approximate indicators of scale rather than directly comparable quantities. We therefore additionally compare language coverage, data type, romanized support, and whether the underlying corpus and construction pipeline are publicly available. Under this comparison, our release contains 2,826B Indic tokens across 22 languages and is, to the best of our knowledge, the largest openly released Indic pretraining corpus reported to date.

\newcolumntype{L}[1]{>{\raggedright\arraybackslash}m{#1}}
\newcolumntype{C}[1]{>{\centering\arraybackslash}m{#1}}
\newcolumntype{R}[1]{>{\raggedleft\arraybackslash}m{#1}}
\newcolumntype{Y}{>{\raggedright\arraybackslash}X} %

\begin{table}[t]
\centering
\small
\setlength{\tabcolsep}{4pt}
\caption{Comparison of Indic pretraining corpora. \# Toks (B): billions of Indic-language tokens as per each source's own tokenizer and reports. Type: Native is human-authored text (web, books, OCR, transcripts); Synth is machine-generated or machine-translated text; Mixed combines both. Roman: whether the corpus includes romanized Indic text. Data Open?: whether the corpus is publicly downloadable. Pipeline Open?: whether the code used to collect, filter or generate the data is public. \cmark yes; \xmark no; \pmark partially released.}
\label{tab:3.3}
\begin{tabular}{L{2.6cm} R{1.2cm} R{1.2cm} L{1.2cm} C{1.6cm} C{1.4cm} C{1.4cm}}
\toprule
\textbf{Corpus} & \textbf{\# Toks (B)} & \textbf{\# Indic Langs} & \textbf{Type} & \textbf{Roman} & \textbf{Data Open?} & \textbf{Pipeline Open?} \\
\midrule
Sangraha~\cite{khan2024indicllmsuite}         & 251       & 22   & Mixed  & \cmark & \cmark  & \cmark \\
BhashaKritika~\cite{manoj2025bhashakritika}    & 540       & 10   & Synth  & --     & \pmark  & \xmark \\
IndicCorp-v2~\cite{doddapaneni2023indiccorp}     & 20.9      & 23   & Native & \xmark & \cmark  & \xmark \\
FineWeb2~\cite{penedo2025fineweb2} & --      & 22 & Native & \xmark & \cmark  & \cmark \\
Sarvam-1~\cite{sarvam2024sarvam1}         & 2,000     & 10   & \xmark     & \xmark     & \xmark  & \xmark \\
Param-1~\cite{pundalik2025param1}          & 2,770 & 1    & Mixed  & --     & \xmark  & \xmark \\
Param-2~\cite{bharatgen2026param2}           & 2,624     & 22   & Native & --     & \xmark  & \xmark \\
\midrule
\textbf{Ours}    & 2,826      & 22   & Synth  & \cmark & \cmark & \cmark \\
\bottomrule
\end{tabular}
\end{table}

Broader multilingual efforts address related challenges in large-scale data collection and curation. FineWeb2 and FinePDFs adapt web processing to individual language--script pairs and study the effects of these choices through training ablations, while HPLT 3.0 combines large-scale web resources with provenance metadata, multilingual evaluation, and model training~\cite{penedo2025fineweb2, oepen2026hplt3}. These efforts reinforce the importance of looking beyond aggregate token counts when characterizing multilingual corpora, including language and script coverage, provenance, transformation history, and the fraction of processed data ultimately released.

\paragraph{Translated and synthetic pretraining data}\label{sec:2.2}
Translation is a common strategy for addressing target-language data scarcity. Doshi et al. study machine-translated Indic documents, target-language filtering, clean-data admixture, and continued pretraining of larger models \cite{doshi2024translationese}. BhashaKritika explores document-, topic-, and persona-grounded Indic generation alongside translation, quality filtering, human assessment, and downstream evaluation \cite{manoj2025bhashakritika}. The MILA dataset similarly presents an integrated Indic corpus effort spanning natural-text acquisition, generation, translation with refinement, human validation, and continued-pretraining experiments~\cite{sawarkar2026mila}. These approaches serve complementary purposes: translation can transfer information contained in selected source documents, while grounded generation can introduce variation in style, context, and content. Neither synthetic route, however, guarantees linguistic naturalness, factual reliability, or representative cultural coverage~\cite{shumailov2024collapse}. At a broader multilingual scale, TransWebEdu translates high-quality English educational web text and evaluates the resulting corpus through language-model training \cite{wang2025transwebedu}. MultiSynt/MT is a particularly relevant recent comparator: it translates Nemotron-CC into 36 languages, preserves source identifiers and aligned translations, compares translation systems, and studies both training utility and limitations in existing evaluation coverage \cite{idahl2026multisynt}. Our work focuses on the Indic-specific setting, with particular attention to source selection, lower-resource languages, script variation, quality-driven repair, and downstream training utility.

\paragraph{Data curation and targeted repair}\label{sec:2.3}

Data-centric pretraining research increasingly treats corpus construction choices as experimental variables. FineWeb and DataComp-LM compare filtering, deduplication, and data selection under controlled training settings \cite{penedo2024fineweb, li2024dclm}, while QuRating learns multidimensional quality scores from LLM preferences and studies their use for data selection and curriculum design \cite{wettig2024qurating}. Nemotron-CC combines classifier-based curation with synthetic rephrasing to investigate the trade-off between data quality and quantity \cite{su2025nemotroncc}. Indic-focused efforts such as Sangraha and BhashaKritika similarly incorporate language-aware processing and explicit quality controls \cite{khan2024indicllmsuite, manoj2025bhashakritika}. For targeted data repair, TEaR uses translation-quality estimates to identify likely errors and guide selective refinement, with evaluation based on automatic metrics and MQM assessments \cite{feng2025tear}. Our pipeline applies quality-driven repair within a source-traceable Indic corpus workflow, with separate checks for language, script, structural preservation, and semantic fidelity. 

\paragraph{Indic evaluation and benchmark reliability}\label{sec:2.4}

Indic evaluation spans several distinct capability dimensions. IndicXTREME focuses on language understanding \cite{doddapaneni2023indiccorp}, IndicGenBench broadens coverage to generation tasks \cite{singh2024indicgenbench}, MILU emphasizes India-centric knowledge \cite{verma2025milu}, and IndicIFEval evaluates instruction following through localized and grounded tasks \cite{jayakumar2026indicifeval}. These benchmarks capture complementary aspects of Indic capability and should not be treated as interchangeable measures of overall model quality. Global MMLU studies cultural sensitivity in multilingual evaluation \cite{singh2025globalmmlu}, while INCLUDE draws on locally sourced examinations across languages and regions \cite{romanou2025include}. Together, these efforts highlight an important distinction between translating a shared capability benchmark into another language and constructing evaluations that measure locally grounded knowledge. 

Benchmark reliability presents an additional challenge. MMLU-Redux documents defects in widely used English evaluations \cite{gema2025mmlu}, while work on EU20 identifies translation and source-level errors in multilingual benchmarks and studies their effect on measured performance \cite{thellmann2024eu20,thellmann2026translationerrors,thellmann2026diagnosing,plaza2024mmlulost}. PARIKSHA similarly shows that human--LLM evaluator agreement can vary across Indic languages and settings \cite{watts2024pariksha}. Together, these findings motivate careful benchmark auditing, versioning, and validation, particularly when adapting evaluation resources to Indic languages.

\paragraph{Our Contributions}\label{sec:2.5}

This work brings these strands together through three connected components: large-scale synthetic Indic pretraining data, a broad and reliability-oriented Indic evaluation suite, and a reusable data-generation pipeline. Relative to prior Indic and multilingual efforts, our emphasis is on jointly studying corpus scale and composition, quality-driven repair, benchmark reliability, and downstream continued-pretraining utility within a source-traceable workflow. We report language and script coverage separately, distinguish generated from retained and released data, and make the processing and validation pipeline reproducible. The resulting model checkpoints are used primarily as experimental instruments for measuring corpus utility and capability retention rather than as the basis for a state-of-the-art model claim.

%% file: paper/03_scope.tex
\section{Scope and Resource Overview}\label{sec:3}

We release four complementary resources: (1) a large-scale Indic pretraining corpus spanning all 22 languages listed in the Eighth Schedule of the Indian Constitution, (2) the pipeline used to construct and quality-control this corpus, (3) the \emph{Indic-LLM-Eval-Suite} for evaluating Indic language models across a broad set of capabilities, and (4) a benchmark construction and validation pipeline for extending reliable evaluation to Indic languages. This section provides an overview of these resources and points to the corresponding sections of the paper.

\subsection{Indic Pretraining Corpus}\label{sec:3.1}

\begin{figure}[t]
  \centering
  \includegraphics[width=\linewidth]{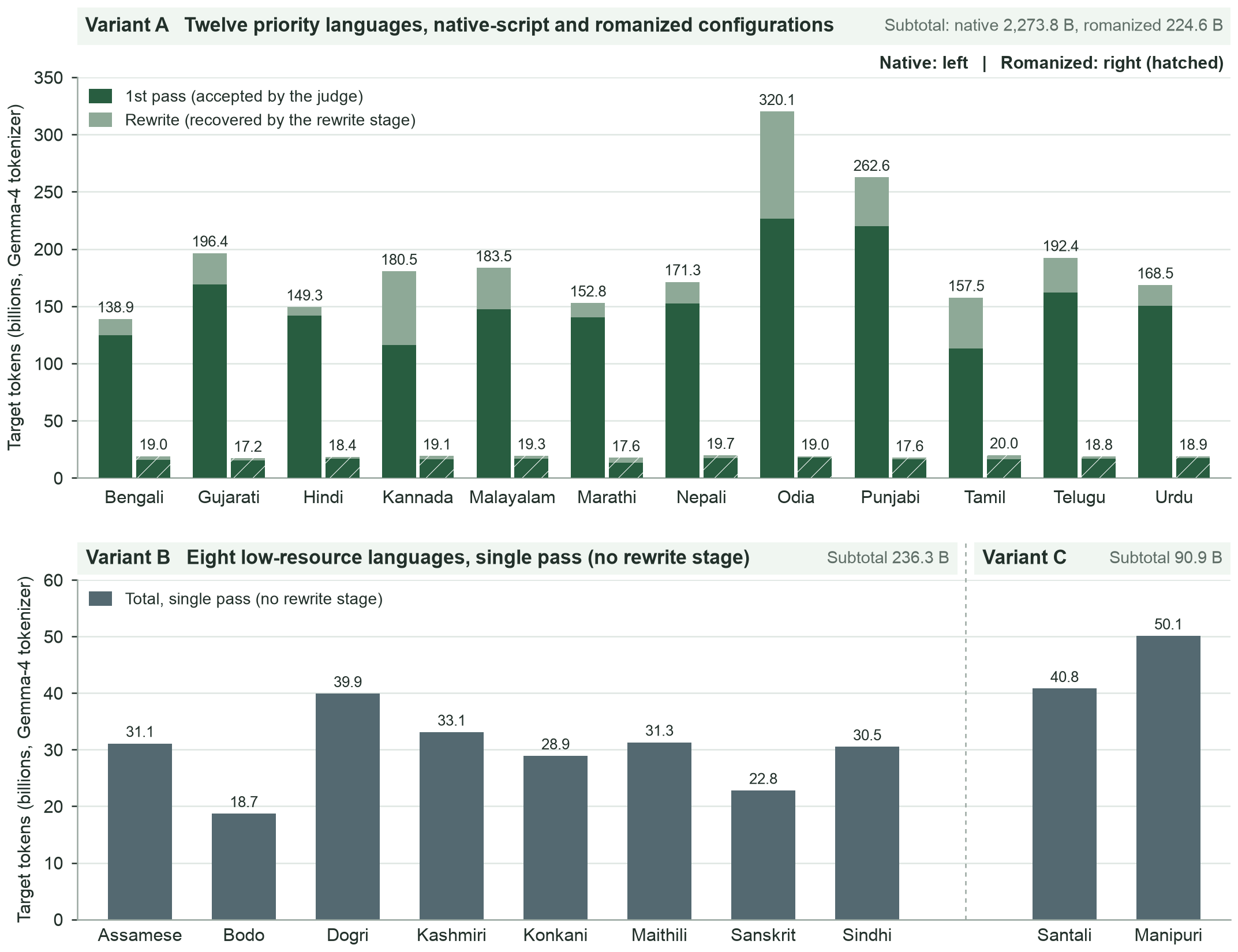}
  \caption{Released target tokens per language, complementing Table~\ref{tab:3.1}. Bars report Gemma-4 target tokens in billions. For the twelve priority languages (Variant A), each native-script bar (left) and romanized bar (right, hatched) is split into tokens accepted directly by the quality judge (first pass) and tokens recovered through the rewriting stage. Variants B and C have no rewrite stage, so only totals are shown. The two rows use different vertical scales.}
  \label{fig:corpus-targets}
\end{figure}

We release pretraining data for all 22 languages listed in the Eighth Schedule of the Indian Constitution, covering 34 language--script configurations. In total, the release contains 2.6T native-script tokens and 225B romanized tokens, measured using the Gemma-4 tokenizer, across 2.44B documents.

Table~\ref{tab:3.1} summarizes the languages, scripts, release configurations, and corpus sizes, and Figure~\ref{fig:corpus-targets} visualizes the resulting target tokens per language. For twelve high-resource Indic languages, we release both a native-script configuration (\textbf{\NPIN{}}) and a romanized configuration in Latin script (\textbf{\NPIR{}}). The remaining ten languages, which are largely low-resource, currently have native-script data only, with romanized configurations planned for subsequent releases. For each dual-script language, the romanized configuration is obtained by transliterating a subset of the accepted native-script translations, corresponding to approximately one tenth of the native-script volume. 

We name each configuration using BCP-47 language, script, and region tags. For example, \texttt{hi-Deva-IN} denotes Hindi in Devanagari script, while \texttt{hi-Latn-IN} denotes its Latin-script romanized configuration. For languages commonly written in multiple scripts, we select a single officially recognized Indic script for the native-script release; in particular, we release Kashmiri in Perso-Arabic script and Sindhi in Devanagari. In Table~\ref{tab:3.1}, \emph{Source} reports the number of English tokens corresponding to the released documents, while \emph{Target} reports the translated token counts.  We describe the corpus counting methodology in Appendix~\ref{app:corpus-accounting}.

All data in the corpus is translated from English documents drawn from Nemotron-CC v2~\cite{su2025nemotroncc}. We sample documents from 15 subsets according to the distribution shown in Table~\ref{tab:3.2}, spanning diverse question-answer data, high-quality organic and rephrased web text, mathematics, code, instruction-style data, and other specialized subsets. Approximately two-fifths of the sampling target comes from the diverse question-answer subset synthesized from Common Crawl. We describe the sampling procedure and source mixture in detail in Section~\ref{sec:bharat-sdg-source}. Every translated record additionally retains an identifier linking it to its corresponding English source document and Nemotron-CC v2 subset, enabling source-level traceability.

\begin{table}[]
\centering
\caption{Overview of the released pretraining corpus by language--script configuration. \emph{Source} reports the number of English tokens corresponding to the released documents, while \emph{Target} reports their translated token counts using the Gemma-4 tokenizer. \emph{1st pass} denotes examples accepted directly by the quality judge, and \emph{Rewrite} denotes examples recovered through the rewriting stage. \emph{Exp.} is the ratio of target to source tokens, and \emph{Docs} denotes the number of released documents. Romanized (\texttt{Latn}) configurations are transliterations of subsets of the corresponding native-script data and therefore provide additional script coverage rather than additional content. We describe this corpus counting methodology in more detail in Appendix~\ref{app:corpus-accounting}.}
\label{tab:3.1}
\scriptsize
\setlength{\tabcolsep}{2.2pt}
\begin{tabular*}{\linewidth}{@{\extracolsep{\fill}}lll rr rrr r r@{}}
\toprule
 & & & \multicolumn{2}{c}{\textbf{Source (English)}} & \multicolumn{3}{c}{\textbf{Target (Gemma-4 tokens, B)}} & & \\
\cmidrule(lr){4-5}\cmidrule(lr){6-8}
\textbf{Language} & \textbf{Config} & \textbf{Script} & Tokens (B) & 1st pass (\%) & 1st pass & Rewrite & Total & \textbf{Exp.} & \textbf{Docs (M)} \\
\midrule
\multicolumn{10}{@{}l}{\textit{Variant A: Gemma-4-31B-it pipeline (Sections~\ref{sec:bharat-sdg-source} to~\ref{sec:bharat-sdg-romanization}); native-script and romanized configurations}} \\
Bengali & \texttt{bn-Beng-IN} & Bengali & 114.0 & 88 & 124.9 & 14.0 & 138.9 & 1.22 & 166.3 \\
 & \texttt{bn-Latn-IN} & Latin & 12.2 & 84 & 15.6 & 3.4 & 19.0 & 1.56 & 16.7 \\
Gujarati & \texttt{gu-Gujr-IN} & Gujarati & 115.5 & 83 & 168.8 & 27.6 & 196.4 & 1.70 & 169.3 \\
 & \texttt{gu-Latn-IN} & Latin & 12.0 & 89 & 15.2 & 2.0 & 17.2 & 1.43 & 16.5 \\
Hindi & \texttt{hi-Deva-IN} & Devanagari & 118.1 & 93 & 142.0 & 7.3 & 149.3 & 1.26 & 171.8 \\
 & \texttt{hi-Latn-IN} & Latin & 12.8 & 91 & 16.9 & 1.5 & 18.4 & 1.44 & 17.2 \\
Kannada & \texttt{kn-Knda-IN} & Kannada & 103.5 & 57 & 116.2 & 64.3 & 180.5 & 1.74 & 156.3 \\
 & \texttt{kn-Latn-IN} & Latin & 11.3 & 82 & 16.4 & 2.7 & 19.1 & 1.69 & 16.7 \\
Malayalam & \texttt{ml-Mlym-IN} & Malayalam & 108.9 & 76 & 147.2 & 36.3 & 183.5 & 1.69 & 161.4 \\
 & \texttt{ml-Latn-IN} & Latin & 11.4 & 83 & 16.7 & 2.6 & 19.3 & 1.69 & 16.0 \\
Marathi & \texttt{mr-Deva-IN} & Devanagari & 114.4 & 90 & 140.4 & 12.4 & 152.8 & 1.34 & 167.7 \\
 & \texttt{mr-Latn-IN} & Latin & 11.6 & 68 & 13.1 & 4.5 & 17.6 & 1.52 & 16.3 \\
Nepali & \texttt{ne-Deva-IN} & Devanagari & 118.0 & 87 & 152.3 & 19.0 & 171.3 & 1.45 & 171.9 \\
 & \texttt{ne-Latn-IN} & Latin & 12.4 & 85 & 17.5 & 2.2 & 19.7 & 1.59 & 17.0 \\
Odia & \texttt{or-Orya-IN} & Odia & 109.3 & 66 & 226.7 & 93.4 & 320.1 & 2.93 & 163.6 \\
 & \texttt{or-Latn-IN} & Latin & 12.2 & 95 & 17.7 & 1.3 & 19.0 & 1.56 & 17.1 \\
Punjabi & \texttt{pa-Guru-IN} & Gurmukhi & 112.3 & 81 & 219.8 & 42.8 & 262.6 & 2.34 & 166.0 \\
 & \texttt{pa-Latn-IN} & Latin & 12.0 & 94 & 16.5 & 1.1 & 17.6 & 1.47 & 16.9 \\
Tamil & \texttt{ta-Taml-IN} & Tamil & 110.5 & 67 & 112.9 & 44.6 & 157.5 & 1.43 & 163.6 \\
 & \texttt{ta-Latn-IN} & Latin & 10.9 & 75 & 16.3 & 3.7 & 20.0 & 1.83 & 16.0 \\
Telugu & \texttt{te-Telu-IN} & Telugu & 115.3 & 81 & 161.8 & 30.6 & 192.4 & 1.67 & 168.6 \\
 & \texttt{te-Latn-IN} & Latin & 12.2 & 92 & 17.0 & 1.8 & 18.8 & 1.54 & 16.9 \\
Urdu & \texttt{ur-Arab-IN} & Perso-Arabic & 117.8 & 86 & 150.6 & 17.9 & 168.5 & 1.43 & 171.6 \\
 & \texttt{ur-Latn-IN} & Latin & 12.8 & 91 & 17.5 & 1.4 & 18.9 & 1.48 & 17.3 \\
\addlinespace[2pt]\multicolumn{3}{@{}l}{\textit{Subtotal, 12 native configurations}} & \textbf{1,357.6} & & \textbf{1,863.6} & \textbf{410.2} & \textbf{2,273.8} & \textbf{1.67} & \textbf{1,998.1} \\
\addlinespace[2pt]\multicolumn{3}{@{}l}{\textit{Subtotal, 12 romanized configurations}} & \textbf{143.8} & & \textbf{196.4} & \textbf{28.2} & \textbf{224.6} & \textbf{1.56} & \textbf{200.7} \\
\addlinespace[2pt]\multicolumn{3}{@{}l}{\textit{Subtotal, 12 priority languages, both scripts}} & \textbf{1,501.4} & & \textbf{2,060.0} & \textbf{438.4} & \textbf{2,498.4} & \textbf{1.66} & \textbf{2,198.8} \\
\midrule
\multicolumn{10}{@{}l}{\textit{Variant B: Bodhan-MT translation, Gemma-4-31B-it judge, no rewrite (Section~\ref{sec:bharat-sdg-additional-languages})}} \\
Assamese & \texttt{as-Beng-IN} & Bengali--Assamese & 16.3 & -- & -- & -- & 31.1 & 1.91 & 29.6 \\
Bodo & \texttt{brx-Deva-IN} & Devanagari & 8.6 & -- & -- & -- & 18.7 & 2.17 & 13.7 \\
Dogri & \texttt{doi-Deva-IN} & Devanagari & 23.4 & -- & -- & -- & 39.9 & 1.71 & 37.7 \\
Kashmiri & \texttt{ks-Arab-IN} & Perso-Arabic & 11.4 & -- & -- & -- & 33.1 & 2.90 & 20.2 \\
Konkani & \texttt{gom-Deva-IN} & Devanagari & 16.7 & -- & -- & -- & 28.9 & 1.73 & 28.9 \\
Maithili & \texttt{mai-Deva-IN} & Devanagari & 20.0 & -- & -- & -- & 31.3 & 1.56 & 33.5 \\
Sanskrit & \texttt{sa-Deva-IN} & Devanagari & 12.6 & -- & -- & -- & 22.8 & 1.81 & 23.7 \\
Sindhi & \texttt{sd-Deva-IN} & Devanagari & 14.6 & -- & -- & -- & 30.5 & 2.09 & 27.5 \\
\addlinespace[2pt]\multicolumn{10}{@{}l}{\textit{Variant C: bespoke routes for Santali and Manipuri (Section~\ref{sec:bharat-sdg-additional-languages})}} \\
Santali & \texttt{sat-Olck-IN} & Ol Chiki & 8.3 & -- & -- & -- & 40.8 & 4.92 & 16.1 \\
Manipuri & \texttt{mni-Mtei-IN} & Meitei Mayek\textsuperscript{a} & 6.6 & -- & -- & -- & 50.1 & 7.59 & 11.2 \\
\addlinespace[2pt]\multicolumn{3}{@{}l}{\textit{Subtotal, 10 low-resource configurations}} & \textbf{138.5} & &  &  & \textbf{327.2} & \textbf{2.36} & \textbf{241.9} \\
\midrule
\multicolumn{3}{@{}l}{\textbf{Total, 34 configurations}} & \textbf{1,639.9} & & \textbf{2,060.0} & \textbf{438.4} & \textbf{2,825.6} & \textbf{1.72} & \textbf{2,440.8} \\
\bottomrule
\end{tabular*}
\smallskip
\raggedright\scriptsize
\textsuperscript{a}\,Translated into Bengali script, then transliterated to Meitei Mayek by rule (Section~\ref{sec:bharat-sdg-additional-languages}). Config names are the released dataset configurations (BCP-47 language-script-region).
\end{table}

\begin{table}[t]
\centering
\caption{Nemotron-CC v2 subsets sampled for translation and their shares of the sampling target (percent). Names are the subset identifiers of the Nemotron-CC v2 release. The retrieval-style question-answer subset \texttt{specialized\_v1\_rqa} (1.00 percent in the initial plan) was dropped before sampling and the remaining shares renormalised.}\label{tab:3.2}
\footnotesize
\begin{tabular}{lr @{\hspace{2.5em}} lr}
\toprule
\textbf{Subset} & \textbf{Share (\%)} & \textbf{Subset} & \textbf{Share (\%)} \\
\midrule
\texttt{cc\_v2\_diverse\_qa} & 40.40 & \texttt{specialized\_v1\_math\_textbooks} & 2.97 \\
\texttt{pretraining\_sft\_v1} & 12.03 & \texttt{specialized\_v1\_wiki\_rewrite} & 2.68 \\
\texttt{cc\_v2\_english\_cc} & 9.56 & \texttt{specialized\_v1\_stem\_sft} & 1.61 \\
\texttt{cc\_math\_v1\_4plus} & 8.65 & \texttt{specialized\_v11\_multiple\_choice} & 1.08 \\
\texttt{cc\_v21\_hq\_synthetic} & 7.46 & \texttt{specialized\_v11\_formal\_logic} & 0.15 \\
\texttt{cc\_v21\_hq\_organic} & 5.59 & \texttt{specialized\_v11\_economics} & 0.10 \\
\texttt{specialized\_v11\_code\_concepts} & 4.67 & \texttt{specialized\_v11\_unconditional\_algo} & 0.06 \\
\texttt{cc\_v21\_hq\_dqa} & 2.99 & & \\
\bottomrule
\end{tabular}
\end{table}

\subsection{Data Construction Pipeline}\label{sec:3.2}

Along with the corpus, we release the pipeline used to construct it. It comprises source-data sampling, large-scale translation, quality assessment, rewriting of rejected examples where applicable, romanization, and final filtering and deduplication. Section~\ref{sec:bharat-sdg-pipeline} describes the pipeline in detail.

We use three pipeline variants to accommodate differences in language coverage and available translation systems. Variant~A (Sections~\ref{sec:bharat-sdg-translation} and~\ref{sec:bharat-sdg-quality}) covers the twelve high-resource languages and uses the gemma-4-31B-it\footnote{\url{https://huggingface.co/google/gemma-4-31b-it}} model for translation, quality assessment, and rewriting. Variant~B (Section~\ref{sec:bharat-sdg-additional-languages}) covers eight additional languages, using the indic-translate\footnote{\url{https://huggingface.co/bodhan-ai/indic-translate}} model for translation and gemma-4-31B-it for quality assessment, with language-specific acceptance thresholds and no rewriting. Variant~C (Section~\ref{sec:bharat-sdg-additional-languages}) handles Santali and Manipuri through specialized language-specific pipelines. These variants correspond directly to the three groups shown in Table~\ref{tab:3.1}, allowing the corpus statistics to be interpreted together with the construction procedure used for each language. Across all variants, the pipeline records source provenance and intermediate quality-control decisions, enabling translated examples to be traced back to their source documents and making the construction process auditable.

\subsection{Indic-LLM-Eval-Suite}\label{sec:3.3}

Alongside the pretraining corpus, we release the \emph{Indic-LLM-Eval-Suite}, an ongoing effort to build a broad and reliable evaluation suite for Indic LLMs. Rather than relying on a small set of aggregate academic benchmarks, the current release evaluates general chat, instruction following, mathematical reasoning, STEM, commonsense reasoning, knowledge, code generation, and translation. Table~\ref{tab:taxonomy} summarizes the capability taxonomy and included benchmarks.

The suite combines existing Indic benchmarks with English benchmarks that can be meaningfully extended through careful translation. However, translation is not appropriate for every capability: evaluations involving local knowledge, cultural context, tools, environments, or interaction settings may require localized or natively constructed benchmarks. Section~\ref{sec:6.2} discusses this distinction and motivates our benchmark selection. All benchmarks are integrated into \textbf{NeMo-Gym}, which provides a common interface for evaluation while preserving benchmark-specific protocols. We release these implementations with the benchmark data to support reproducible and extensible evaluation. In Section~\ref{sec:6.5}, we use this setup to evaluate contemporary open-weight Indic and multilingual models across capabilities and languages.

We view the current suite as a versioned and evolving resource rather than a fixed benchmark collection. As existing evaluations saturate, new model capabilities emerge, and better Indic-specific benchmarks become available, we plan to continuously expand and revise the suite.

\subsection{Benchmark Construction and Validation Pipeline}\label{sec:3.4}

A central component of the evaluation release is the pipeline used to translate, correct, and validate benchmarks before their inclusion in the suite. Our goal is not only to expand multilingual benchmark coverage, but also to ensure that the resulting scores continue to measure the capabilities intended by the original evaluations. We discuss the challenges underlying this process in Section~\ref{sec:6.1} and describe the complete construction and validation procedure in Sections~\ref{sec:6.3} \&~\ref{sec:6.4}.

For benchmarks that require translation, we use a benchmark-aware human-in-the-loop pipeline rather than a generic translation prompt. This pipeline comprises \emph{translate $\rightarrow$ evaluate $\rightarrow$ correct $\rightarrow$ evaluate} stages, which we describe in detail in Section~\ref{sec:6.3}. We then ask native speakers to independently validate the benchmark instances. Section~\ref{sec:6.4} describes the details of the annotation process. Together, these components provide a common framework for constructing and maintaining translated benchmarks in the \emph{Indic-LLM-Eval-Suite}, while preserving benchmark-specific task requirements and keeping intermediate quality-control decisions auditable.

%% file: paper/04_pipeline.tex
\section{The Indic Synthetic Data Pipeline}
\label{sec:bharat-sdg-pipeline}

We construct the pretraining corpus introduced in Section~3.1 by translating curated English documents. We check each translation against its source, repair low-quality outputs where applicable, and retain the source and quality metadata for each accepted record.

We use \href{https://docs.nvidia.com/nemo/datadesigner/}{NeMo Data Designer} to run translation and quality assessment. \href{https://huggingface.co/google/gemma-4-31B-it}{Gemma-4-31B-it} translates the top-12 Indic languages, for which we generate native-script and romanized data. \href{https://huggingface.co/bodhan-ai/indic-translate}{Bodhan's Indic Translate model}, referred to here as Bodhan-MT, translates the remaining ten Indic languages, for which we generate native-script data only. Figures~\ref{fig:bharat-sdg-priority} and~\ref{fig:bharat-sdg-additional} show these training-data workflows, including their quality checks and final outputs. Section~6 describes the separate pipeline used to construct and validate evaluation benchmarks.

\subsection{Source selection and document preparation}
\label{sec:bharat-sdg-source}

We sample English documents from the source subsets listed in Table~3.2 to cover several domains and text formats. We retain source identifiers throughout generation so that we can trace each output to its input and check the resulting source distribution.

For the top-12 workflow, we select documents with fewer than 4,096 Gemma-4 tokens and translate each document in full, without splitting. Keeping the document together gives the translator access to the context around a paragraph, equation, table, or code block. Workflows for the remaining ten languages use the preparation described in Section~\ref{sec:bharat-sdg-additional-languages}, including a separate windowed path for Santali.

The translation instructions distinguish prose from content that should remain unchanged. We ask the model to preserve facts, references to people, places, and organizations, document order, and formatting, while leaving code, mathematical expressions, URLs, and file paths intact. Quality checks then assess whether the translation follows these instructions.

\subsection{Translation for the top-12 Indic languages}
\label{sec:bharat-sdg-translation}

The top-12 Indic languages are Bengali, Gujarati, Hindi, Kannada, Malayalam, Marathi, Nepali, Odia, Punjabi, Tamil, Telugu, and Urdu. For these languages, we translate each selected document in full. The model must therefore preserve the document's organization and technical content, not just translate individual sentences.

To choose the translator, we compared several models, including Gemma-4 variants, IndicTrans2, and Sarvam-translate, on sentence benchmarks and complete documents. We also checked how well they preserved code, equations, tables, and other structured content. Specialist translation models performed strongly on sentence benchmarks, but those scores did not consistently predict their handling of complete documents. These document-level comparisons led us to select Gemma-4-31B-it for the top-12 workflow. Appendix C lists the candidate models, their decoding settings, and their sentence-level scores; the document-level comparisons were development studies and are not reported in this version.

To preserve the document's structure, we request a plain-text translation rather than a JSON response. The prompt asks for the translated document itself, without introductory remarks or explanations. This avoids adding an output-format requirement that can interfere with mathematics, LaTeX, or other structured text.

\begin{figure}[htbp]
  \centering
  \includegraphics[width=\linewidth]{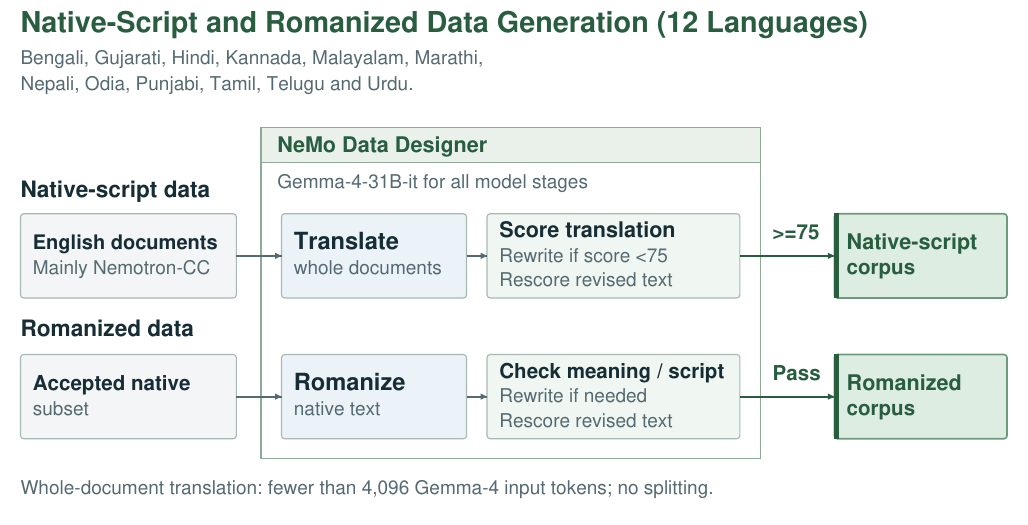}
  \caption{Workflow for the top-12 Indic languages (Variant A in Section~3.2). Gemma-4-31B-it translates whole English documents with fewer than 4,096 Gemma-4 tokens, assesses the translations, and rewrites low-scoring candidates before reassessment. Corpus assembly retains native translations that meet the score threshold of 75. A subset of the retained native-script text is romanized and checked against its native-script input. NeMo Data Designer runs generation and assessment; corpus acceptance follows these steps. Appendix~\ref{app:bharat-sdg-execution} describes inference optimizations and execution details.}
  \label{fig:bharat-sdg-priority}
\end{figure}

The prompts also distinguish numerals in ordinary prose from digits in technical content. Where a configuration requests native numerals in prose, it still requires ASCII digits inside code, formulas, and URLs. We record the requested numeral style with each document. Figure~\ref{fig:bharat-sdg-priority} summarizes the translation, assessment, and romanization steps for these languages.

\subsection{Quality assessment and targeted rewriting}
\label{sec:bharat-sdg-quality}

For the top-12 workflow, Gemma-4-31B-it also acts as the judge: it compares each candidate with its English source. We use a rubric inspired by Multidimensional Quality Metrics (MQM) to assess meaning, fluency, terminology, style, and format preservation. The judge returns a score from 0 to 100, identifies the affected text and error severity, and briefly explains its assessment. The rewrite stage uses this feedback to address specific problems.

We retain first-pass translations with a score of at least 75. For lower-scoring candidates, we provide the source, translation, and judge feedback to the rewrite stage. The model returns a corrected translation, which we assess again. We retain the rewritten translation if its reassessed score reaches 75; otherwise, we exclude it. This policy applies to the top-12 workflow. The remaining ten languages use the policies described in Section~\ref{sec:bharat-sdg-additional-languages}.

Rewriting lets us recover documents that would otherwise be discarded after the first assessment. For corpus accounting, we distinguish first-pass acceptance, recovery after rewriting, and final rejection. Retained corpus counts therefore differ from the number of generated attempts. Table~3.1 separates retained first-pass and rewritten tokens for the configurations that use rewriting; Appendix~B explains the accounting conventions.

We complement automated scoring with sampled human review. An initial review covered 92 Hindi documents, and subsequent reviews identified residual English words and inconsistent handling of technical terms. These checks help us understand the errors that remain in generated text; their coverage is not yet comparable across languages. Because the translator also acts as the judge, possible self-preference remains a consideration when interpreting its scores.

To test whether the judge detects specific problems, we also evaluated translations with deliberately introduced errors: changed names or numbers, omitted information, reversed meaning, and unwanted language mixing. These tests identified omissions and residual English as the judge's weakest error categories and informed refinements to the scoring prompts. The judge's error descriptions guide the rewrite stage to restore missing content and correct untranslated or mistranslated spans; reassessment checks whether the revised output meets the acceptance criterion. Details regarding the prompts and their release are provided in Appendix~\ref{sec:A}. Appendix~\ref{app:analysis} reports development analyses of two recurring failure modes, numeral preservation and script contamination.

\subsection{Romanized data}
\label{sec:bharat-sdg-romanization}

We create romanized data by converting a subset of the native-script translations for the top-12 Indic languages into Latin script. This two-step path separates translation from script conversion: the first stage translates the English document, and the second changes how the Indic text is written while preserving its meaning.

The romanization stage uses Gemma-4-31B-it with script-specific instructions. We compare the romanized candidate with its native-script input, checking meaning and use of the requested script, and use feedback-based rewriting where needed. Each record also retains the identifier of its English source.

In our initial human comparison, direct English-to-romanized generation produced more hallucinated content and residual English than the two-step route. We therefore selected native-script translation followed by transliteration.

Romanized records express a subset of the native corpus in another script; they are not an independent collection of source documents. Table~3.1 reports native and romanized token volumes separately and includes both in the corpus total.

\subsection{Translation for the remaining ten Indic languages}
\label{sec:bharat-sdg-additional-languages}

\begin{figure}[htbp]
  \centering
  \includegraphics[width=\linewidth]{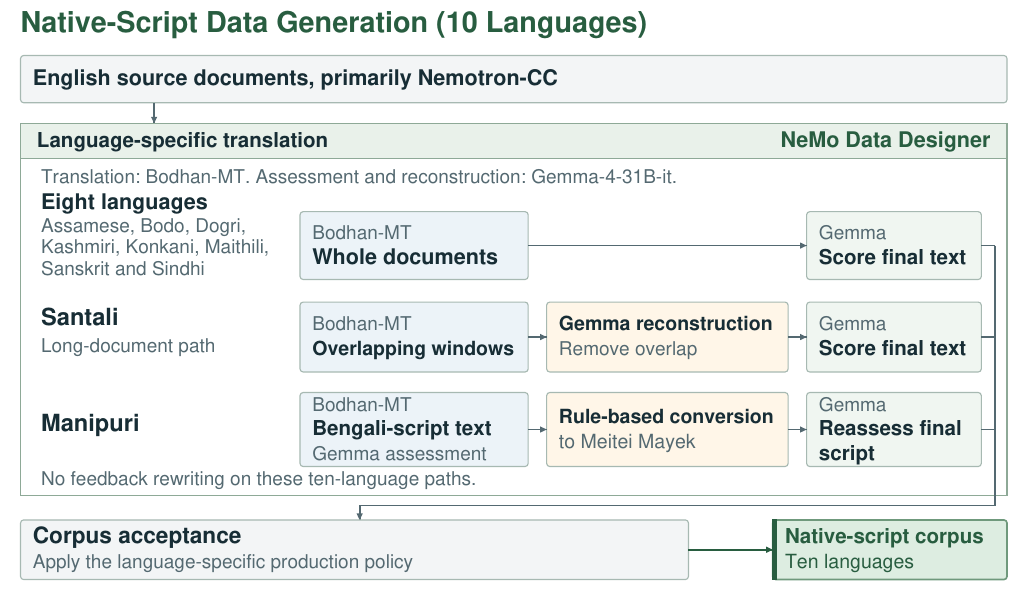}
  \caption{Workflows for the remaining ten Indic languages (Variants B and C in Section~3.2), using Bodhan-MT for translation and Gemma-4-31B-it for quality assessment. Eight languages use whole-document translation. Longer Santali documents use overlapping windows followed by reconstruction. Manipuri uses Bengali-script translation followed by conversion to Meitei Mayek, with assessment before and after conversion. Corpus assembly applies language-specific acceptance rules; these workflows do not use the top-12 feedback-rewrite loop. Appendix~\ref{app:bharat-sdg-execution} describes inference optimizations and execution details.}
  \label{fig:bharat-sdg-additional}
\end{figure}

The remaining ten Indic languages are Assamese, Bodo, Dogri, Kashmiri, Konkani, Maithili, Manipuri, Sanskrit, Santali, and Sindhi. Our comparisons found that translation quality and generation stability varied with language, document length, and script. We use Bodhan-MT to translate these languages, with separate handling of longer Santali documents and Manipuri script conversion. Figure~\ref{fig:bharat-sdg-additional} shows these three approaches. All retain the English source identifiers and assess translations against their sources.

For the eight languages other than Santali and Manipuri, Bodhan-MT translates whole documents. Gemma-4-31B-it then assesses each output against its English source. These workflows use language-specific acceptance thresholds and do not use the top-12 rewrite loop.

For longer Santali documents, we divide the source at line breaks into overlapping windows and translate each window with Bodhan-MT. These initial calls do not receive earlier translations. Gemma-4-31B-it then uses neighboring source windows and their translations to select the text for each non-overlapping source segment and make limited consistency corrections. We join these segments in document order and assess the complete output. This uses local context during reconstruction while avoiding repeated text from the overlapping windows.

For Manipuri, Bodhan-MT first produces a Manipuri translation written in Bengali script. We then apply a fixed character-mapping table to convert it to Meitei Mayek, because direct long-form generation in Meitei Mayek was less stable in our comparisons. Gemma-4-31B-it assesses the Bengali-script translation and reassesses the converted output against the English source. Some Bengali characters map to the same Meitei character, and some outputs retain names in Latin script. These cases still need native-reader validation.

We compared these approaches across document lengths, checking translation quality and preservation of code, tables, mathematics, and other structure. The reported studies have not yet established how well judge scores for the remaining ten languages agree with native-speaker assessments. We report this validation status separately from the Hindi pilot and subsequent reviews described in Section~\ref{sec:bharat-sdg-quality}.

After generation, corpus assembly applies each workflow's acceptance policy and retains source and quality metadata (Appendix~\ref{app:bharat-sdg-execution}). Section~5 evaluates first-pass accepted subsets of the data for the top-12 Indic languages in controlled continued-pretraining (CPT) experiments. These experiments began before rewriting was complete; the full release also includes data recovered through rewriting and data for the remaining ten Indic languages.

%% file: paper/06_experiments.tex
\providecommand{\BharatCorpusName}{Nemotron-Pretraining-Indic-v1}
\providecommand{\BharatCorpusSection}{Section~3.1}
\providecommand{\BharatPipelineSection}{Section~4}
\providecommand{\BharatEvaluationSection}{Section~6}
\providecommand{\BharatSangrahaRef}{\cite{khan2024indicllmsuite}}
\providecommand{\BharatMILURef}{\cite{verma2025milu}}

\section{Controlled Continued-Pretraining Experiments}
\label{sec:bharat-cpt-experiments}

We evaluate how \BharatCorpusName{} (\NPIv{}) complements natural Indic text during continued pretraining. Using this synthetic corpus effectively requires deciding which data to include, which scripts to cover, and how much of the training budget to allocate to Indic languages. The experiments examine these choices through comparisons of data sources, synthetic-data volume, and training mixtures, with English benchmarks measuring the accompanying changes in broader capabilities.

The study focuses on first-pass accepted translations from \BharatPipelineSection{} for the top-12 Indic languages: Bengali, Gujarati, Hindi, Kannada, Malayalam, Marathi, Nepali, Odia, Punjabi, Tamil, Telugu, and Urdu. This twelve-language subset defines the scope of the results below. The full release also includes translations recovered through rewriting and data for the remaining ten languages.

\subsection{Experimental setup}
\label{sec:bharat-cpt-setup}

The experiments use Nemotron Nano models during the learning-rate decay phase of pretraining. The first set of comparisons asks which Indic data to use: each run processes 100 billion (100B) tokens, including 45B Indic tokens, while varying the data sources, script mixture, or available synthetic-data pool. The common Indic allocation provides a consistent scale for comparing these data choices within each study.

The remaining 55B tokens come from the original Nemotron non-Indic pretraining mixture, which contains English web text, code, mathematics, synthetic text, and task data.\footnote{Publicly released Nemotron pretraining data are listed in NVIDIA's \href{https://huggingface.co/collections/nvidia/nemotron-pre-training-datasets}{Nemotron Pre-Training Datasets collection}. This collection is a public resource reference, not the exact sampling recipe for these experiments.} A second set of comparisons asks how much Indic data to include overall. These runs use a 200B-token budget and vary the Indic share from 10\% to 30\%, assessing Indic and English performance together. Table~\ref{tab:bharat-cpt-design} summarizes both sets of experiments.

\subsection{Training data and mixtures}
\label{sec:bharat-cpt-data}

The Indic training mixtures draw on natural text and synthetic translations. Sangraha's verified and unverified web splits~\BharatSangrahaRef{}, together with FineWeb2 and FinePDFs, supply the natural-text component: FineWeb2 contributes web text, while FinePDFs contributes text extracted from PDFs. Sangraha's translated split adds an established synthetic-data baseline, produced by translating English Wikimedia text with IndicTrans2. These four publicly available pools contain 352B tokens across 245M documents in total (Table~\ref{tab:bharat-cpt-pools}).

To assess the contribution of \NPIv{}, the subsequent comparisons combine its synthetic data with natural Indic text from Sangraha. The native-script mixture uses first-pass accepted translations, while the mixed-script mixture also includes romanized text. Both forms come from the same synthetic-data pipeline; \emph{native-script} refers to the writing system, rather than text originally authored in an Indic language. Because the translated corpora differ in their English sources, translation models, and quality checks, the results compare complete datasets and mixtures.

Table~\ref{tab:bharat-cpt-pools} reports the text available in each data pool, rather than the tokens consumed by a training run. Training may sample only part of a pool or repeat documents to fill its allocation. Token counts here follow the experimental tokenizer, recorded as \texttt{multiMixV8}; the release inventory in \BharatCorpusSection{} uses Gemma-4. The same text can have different token counts under these tokenizers, so the inventory and training counts should be read separately.

\begin{table}[htbp]
\centering
\small
\caption{Data available for constructing the Indic training mixtures. Counts describe the source pools, before sampling for a training run. \NPIN{} and \NPIR{} identify the native-script and romanized subsets of \NPIv{} (\BharatCorpusSection{}).}
\label{tab:bharat-cpt-pools}
\begin{tabularx}{\linewidth}{@{}l p{0.24\linewidth} X p{0.16\linewidth}@{}}
\toprule
Label & Pool & Description & \shortstack[l]{Available tokens\\(billions)} \\
\midrule
SG-V & Sangraha verified & Web text from human-verified sources & 85.6 \\
SG-U & Sangraha unverified & Perplexity-filtered web text & 32.6 \\
SG-T & Sangraha translated & English Wikimedia translated with IndicTrans2 & 139.0 \\
FW+FP & FineWeb2 + FinePDFs & Web text and PDF-extracted text & 94.8 \\
\NPIN{} & \NPIv{}, native script & First-pass accepted translations, score at least 75 & Varies by study \\
\NPIR{} & \NPIv{}, romanized & First-pass accepted romanized text & Varies by study \\
\bottomrule
\end{tabularx}
\end{table}

\subsection{Evaluation}
\label{sec:bharat-cpt-evaluation}

Four benchmarks assess Indic knowledge and reasoning: MILU~\BharatMILURef{} and translated versions of MMLU-Pro, ARC-Challenge, and GSM8K. The source survey averages MILU over ten Indic languages and English. The multilingual comparison presents means with and without Odia, with the results including Odia treated as primary. The second set of means shows how Odia's lower scores affect the aggregate results; the reasons for those lower scores remain an open question.

The main comparisons concern training configurations within each study. Gemma-4-26B-A4B, abbreviated Gemma-26B in the tables, provides an external reference for benchmark performance rather than a matched training control. Rows labeled Base report the starting Nemotron Nano checkpoint before continued pretraining.

The 200B-token study also measures English performance through twelve metric combinations across ten benchmarks. Its Indic results include both the individual benchmark scores and their arithmetic mean across MILU, MMLU-Pro, ARC-Challenge, and GSM8K. Each configuration has one training run, so small score differences are descriptive observations rather than evidence of statistical equivalence or repeatability.

\BharatEvaluationSection{} describes the separate benchmark audit and validation process. Re-evaluation of these training configurations on the audited V1 suite and MILU 1.5 remains pending. For the 100B-token configurations, further evaluation would require retraining because their checkpoints were not retained.

\subsection{Experiments and results}
\label{sec:bharat-cpt-results}

The experiments address two related questions: which Indic data to include, and how much of the training mixture to allocate to it. The first comparison tests publicly available natural and translated corpora to establish baseline mixtures. The next comparisons combine natural text with \NPIv{}, vary the available synthetic-data pool, and evaluate romanized data. A separate study varies the overall Indic share and measures both Indic gains and changes on English benchmarks. Table~\ref{tab:bharat-cpt-design} summarizes the settings.

\begin{table}[htbp]
\centering
\small
\caption{Experiment overview. Budgets denote sampled training tokens, not available corpus size.}
\label{tab:bharat-cpt-design}
\begin{tabularx}{\linewidth}{@{}X X r r@{}}
\toprule
Comparison & What changes & Total budget & Indic budget \\
\midrule
Public Indic corpora (\ref{sec:bharat-cpt-sources}) & Allocation across natural and translated sources & 100B & 45B \\
\NPIv{} contribution (\ref{sec:bharat-cpt-recipes}) & Mixtures with \NPIv{} data; script mix & 100B & 45B \\
Pool size (\ref{sec:bharat-cpt-scaling}) & Punjabi or Malayalam pool configurations & 100B & 45B \\
Romanized Malayalam (\ref{sec:bharat-cpt-romanized}) & Native-script versus romanized training recipes & 100B & 45B \\
Indic share (\ref{sec:bharat-cpt-share}) & 10\%, 20\%, or 30\% Indic sampling & 200B & 20B/40B/60B \\
English results (\ref{sec:bharat-cpt-english}) & Evaluation of the same Indic-share runs & 200B & 20B/40B/60B \\
\bottomrule
\end{tabularx}
\end{table}

\subsubsection{Publicly available Indic corpora}
\label{sec:bharat-cpt-sources}

The baseline comparison tests five mixtures of publicly available Indic corpora, spanning natural web and PDF text and Sangraha's translated data (Table~\ref{tab:bharat-cpt-sources}). Each run processes 100B tokens, with 45B drawn from the Indic mixture.

The natural-text configurations perform similarly on MILU: FineWeb2/FinePDFs reaches 61.7, Sangraha's verified/unverified web mixture reaches 62.0, and their combination reaches 62.1. Sangraha's translated split scores 60.8 on its own. The mixture containing all four pools achieves the highest score among the tested settings, 63.3.

The results show why standalone performance is only one part of assessing a training source. Although the translated split scores lowest on its own, it also contributes to the best-performing mixture in this comparison. The other source shares change as well, so this result supports the combined mixture rather than isolating the benefit of translated text.

\begin{table}[htbp]
\centering
\small
\caption{Source survey: 100B-token runs with 45B Indic tokens. Recipe numbers are percentages of all training tokens; the remaining 55\% uses the base mixture. MILU is averaged over eleven languages, including English.}
\label{tab:bharat-cpt-sources}
\begin{tabularx}{\linewidth}{@{}X r r@{}}
\toprule
Recipe & Indic tokens (B) & MILU \\
\midrule
Base & -- & 47.1 \\
FW+FP 45 & 45 & 61.7 \\
SG-T 45 & 45 & 60.8 \\
SG-V 33 + SG-U 12 & 45 & 62.0 \\
SG-V 18 + SG-U 6 + FW+FP 21 & 45 & 62.1 \\
SG-V 15 + SG-U 4 + SG-T 8 + FW+FP 18 & 45 & 63.3 \\
Gemma-26B & -- & 65.6 \\
\bottomrule
\end{tabularx}
\end{table}

\subsubsection{Contribution of NPI-v1}
\label{sec:bharat-cpt-recipes}

This comparison tests whether combining \NPIv{} with natural Indic text improves on a mixture built from the public corpora above (Table~\ref{tab:bharat-cpt-recipes}). Both \NPIv{} configurations use first-pass accepted synthetic data alongside Sangraha web text. The native-script configuration assigns 25\% of all training tokens to \NPIN{} data and 20\% to Sangraha: 18\% verified and 2\% unverified. The mixed-script configuration keeps the same Sangraha shares and replaces five percentage points of \NPIN{} data with \NPIR{} data.

Relative to the public-corpora baseline in this comparison, the \NPIN{} mixture improves the reported means by 1.3 points on MILU, 1.6 on MMLU-Pro, 3.4 on ARC-Challenge, and 8.0 on GSM8K. The largest improvement is on GSM8K. For context, the recipe also scores above Gemma-26B on MMLU-Pro and GSM8K, while its MILU and ARC-Challenge scores are close to that reference.

Replacing five percentage points of native-script data with romanized data changes each reported native-script score by no more than 0.6 points. Section~\ref{sec:bharat-cpt-romanized} examines romanized inputs separately.

These results support combining \NPIv{} data with natural Indic text. One possible explanation for the reasoning gains is that Indic training helps the model use capabilities learned from English mathematics and code. Distinguishing this transfer from learning additional task knowledge would require a separate experiment.

\begin{table}[htbp]
\centering
\small
\caption{Mixtures containing \NPIv{} data at a 100B-token budget and 45\% Indic share. The second panel excludes Odia as a sensitivity analysis. Recipe shares are percentages of all training tokens.}
\label{tab:bharat-cpt-recipes}
\begin{tabularx}{\linewidth}{@{}X r r r r@{}}
\toprule
\multicolumn{5}{l}{\textbf{Reported inclusive means}} \\
Recipe & MILU & MMLU-Pro & ARC-C & GSM8K \\
\midrule
Base & 44.4 & 29.3 & 53.7 & 40.4 \\
Public-corpora baseline & 63.2 & 40.2 & 84.2 & 67.7 \\
\NPIN{} mixture & 64.5 & 41.8 & 87.6 & 75.7 \\
\NPIN{} + \NPIR{} mixture & 64.3 & 42.0 & 87.3 & 75.1 \\
Gemma-26B & 64.7 & 34.8 & 87.3 & 61.9 \\
\midrule
\multicolumn{5}{l}{\textbf{Sensitivity analysis: Odia excluded}} \\
Recipe & MILU & MMLU-Pro & ARC-C & GSM8K \\
\midrule
Base & 45.8 & 30.3 & 56.1 & 43.8 \\
Public-corpora baseline & 64.7 & 41.1 & 85.4 & 68.4 \\
\NPIN{} mixture & 65.8 & 42.6 & 88.2 & 75.8 \\
\NPIN{} + \NPIR{} mixture & 65.6 & 42.7 & 87.8 & 75.2 \\
Gemma-26B & 65.2 & 35.0 & 87.7 & 61.8 \\
\bottomrule
\end{tabularx}
\par\smallskip
\footnotesize \NPIN{} mixture: \NPIN{} 25 + SG-V 18 + SG-U 2. \NPIN{} + \NPIR{} mixture: \NPIN{} 20 + \NPIR{} 5 + SG-V 18 + SG-U 2. Public-corpora baseline: SG-V/U + SG-T + FW+FP.
\end{table}

\subsubsection{Size of the synthetic-data pool}
\label{sec:bharat-cpt-scaling}

The next comparison examines how performance changes with the size of the available \NPIv{} synthetic-data pool. Separate Punjabi and Malayalam studies allocate the Indic portion of training to one language at a time (Table~\ref{tab:bharat-cpt-scaling}), using the reported 100B-token budget with 45B Indic tokens. Because source allocation and repetition may also vary between configurations, the results describe these data settings together rather than the effect of unique-data volume alone.

In both languages, the transition from the web-only baseline to the 10B configuration provides the largest observed stepwise improvement. Punjabi GSM8K rises from 66.7 to 78.2, and Malayalam GSM8K rises from 69.2 to 76.5. MILU, MMLU-Pro, and ARC-Challenge also improve.

Larger configurations provide smaller additional gains. From the 10B to the 30B configuration, Punjabi MILU increases from 63.1 to 65.8 and GSM8K from 78.2 to 79.8. For Malayalam, moving from 10B to 50B increases MILU from 62.3 to 64.0 and GSM8K from 76.5 to 78.5. The intermediate results are not uniformly increasing. These observations favor testing pool size together with sampling allocation and repetition when deciding how much translated data to generate and consume.

\begin{table}[htbp]
\centering
\small
\caption{Single-language pool-size comparisons. ``Web only'' contains Indic data but excludes our synthetic data.}
\label{tab:bharat-cpt-scaling}
\begin{tabularx}{\linewidth}{@{}X r r r r@{}}
\toprule
\multicolumn{5}{l}{\textbf{Punjabi}} \\
Pool configuration & MILU & MMLU-Pro & ARC-C & GSM8K \\
\midrule
Web only & 59.06 & 38.05 & 81.45 & 66.72 \\
10B & 63.06 & 42.99 & 87.49 & 78.24 \\
20B & 65.24 & 43.96 & 89.04 & 78.85 \\
30B & 65.80 & 44.30 & 89.04 & 79.83 \\
Gemma-26B & 65.58 & 34.90 & 88.20 & 64.70 \\
\midrule
\multicolumn{5}{l}{\textbf{Malayalam}} \\
Pool configuration & MILU & MMLU-Pro & ARC-C & GSM8K \\
\midrule
Web only & 59.66 & 37.77 & 82.05 & 69.22 \\
10B & 62.25 & 41.80 & 87.92 & 76.50 \\
30B & 63.20 & 41.85 & 88.96 & 78.70 \\
50B & 64.01 & 42.93 & 88.78 & 78.47 \\
Gemma-26B & 63.76 & 35.10 & 88.10 & 62.50 \\
\bottomrule
\end{tabularx}
\end{table}

\subsubsection{Romanized data for Malayalam}
\label{sec:bharat-cpt-romanized}

To examine the role of script coverage, we evaluate the Malayalam configurations on romanized MILU and GSM8K (Table~\ref{tab:bharat-cpt-romanized}). The setup uses a 100B-token training budget with 45B Indic tokens and ASCII-digit scoring for GSM8K.

The native-script configuration with a 50B pool reaches 47.0 on romanized MILU and 30.1 on romanized GSM8K. The romanized-data configuration reaches 64.4 and 73.1, gains of 17.4 and 43.0 points. The romanized recipe performs substantially better on these Malayalam tests. Alongside the small native-script differences in Section~\ref{sec:bharat-cpt-recipes}, this motivates including script coverage explicitly in data selection and evaluation. The recipes also differ in other respects, so the comparison does not isolate the contribution of script alone.

\begin{table}[htbp]
\centering
\small
\caption{Performance on romanized Malayalam MILU and GSM8K. GSM8K uses ASCII-digit scoring.}
\label{tab:bharat-cpt-romanized}
\begin{tabularx}{\linewidth}{@{}X r r@{}}
\toprule
Training recipe & MILU, romanized & GSM8K, romanized \\
\midrule
Base & 44.25 & 19.64 \\
Sangraha web & 45.73 & 21.91 \\
Sangraha + \NPIN{} (50B pool) & 47.00 & 30.10 \\
Sangraha + \NPIR{} & 64.43 & 73.09 \\
Gemma-26B & 56.84 & 63.46 \\
\bottomrule
\end{tabularx}
\end{table}

\subsubsection{Indic share of the training mixture}
\label{sec:bharat-cpt-share}

After comparing the choice of Indic data, this study examines its share of the overall training mixture. Three configurations train for 200B tokens with Indic shares of 10\%, 20\%, and 30\%, corresponding to 20B, 40B, and 60B Indic training tokens (Table~\ref{tab:bharat-cpt-share}). Each setting uses a multilingual Indic mixture, and the same runs support the English evaluation below.

All four Indic benchmark scores increase as the share rises across these settings. The four-benchmark mean increases from 63.1 at 10\% to 64.6 at 20\% and 65.6 at 30\%. Thus, the two increases add approximately 1.5 and 1.0 points, respectively. The 30\% configuration gives the highest Indic scores among the settings tested.

The reference comparison also varies by task: at 30\%, scores exceed Gemma-26B on MMLU-Pro and GSM8K, but remain below it on MILU and ARC-Challenge. The aggregate mean should therefore be read alongside the individual benchmarks. The English results below complete this comparison.

\begin{table}[htbp]
\centering
\small
\caption{Indic-share study: 200B total training tokens. Indic token allocations include the full Indic mixture, not only our synthetic data. Mean-4 is the arithmetic mean of the four displayed benchmark scores.}
\label{tab:bharat-cpt-share}
\begin{tabularx}{\linewidth}{@{}X r r r r r@{}}
\toprule
Configuration & MILU & MMLU-Pro & ARC-C & GSM8K & Mean-4 \\
\midrule
Base & 46.8 & 33.5 & 57.5 & 44.1 & 45.5 \\
Indic 10\% (20B) & 58.8 & 41.9 & 82.2 & 69.5 & 63.1 \\
Indic 20\% (40B) & 60.2 & 43.0 & 83.9 & 71.4 & 64.6 \\
Indic 30\% (60B) & 61.1 & 43.8 & 85.0 & 72.3 & 65.6 \\
Gemma-26B & 64.7 & 34.8 & 87.3 & 61.9 & 62.2 \\
\bottomrule
\end{tabularx}
\end{table}

\subsubsection{English benchmark performance}
\label{sec:bharat-cpt-english}

English evaluation checks how the three Indic allocations affect performance on code, mathematics, commonsense, and truthfulness tasks. Table~\ref{tab:bharat-cpt-english} reports twelve metric combinations across ten benchmarks as changes relative to the base comparator.

The results are mixed across tasks. BigCodeBench improves by 11.7--14.3 points, and several other code and mathematics metrics also increase. Six metrics remain below baseline at all three shares. These include GSM8K, with declines of 0.2--2.4 points, and Minerva MATH-500, with declines of 1.5--2.6 points. No metric declines monotonically as the Indic share increases from 10\% to 30\%.

The Indic gains therefore accompany improvements on some English tasks and declines on others. With one training run per setting, we cannot determine whether the smaller differences reflect the mixture or training and evaluation variability. A training recipe should be selected using the task-level English results alongside Indic performance, rather than the Indic mean alone.

\begin{table}[htbp]
\centering
\small
\caption{English benchmark changes for the 200B-token Indic-share study. Values are percentage-point changes relative to the base comparator.}
\label{tab:bharat-cpt-english}
\begin{tabularx}{\linewidth}{@{}X r r r@{}}
\toprule
Benchmark and metric & 10\% Indic & 20\% Indic & 30\% Indic \\
\midrule
BigCodeBench instruct, pass@1 & +13.4 & +14.3 & +11.7 \\
HMMT, pass@32 & +10.0 & +3.3 & +10.0 \\
LiveCodeBench sampled, pass@32 & +3.0 & +0.1 & +1.5 \\
HMMT, pass@1 & 0.0 & +2.0 & +0.3 \\
MATH-500, pass@1 & +1.3 & +0.9 & +1.6 \\
HumanEval, pass@1 & +0.6 & +1.0 & +0.9 \\
GSM8K, exact match & -2.4 & -1.1 & -0.2 \\
Minerva MATH-500, exact match & -2.1 & -1.5 & -2.6 \\
OpenBookQA, acc\_norm & -1.8 & -0.4 & -0.4 \\
TruthfulQA MC2 & -1.0 & -1.6 & -1.3 \\
CommonsenseQA & -1.4 & -1.7 & -0.3 \\
LiveCodeBench, pass@1 & -0.4 & -1.1 & -0.4 \\
\bottomrule
\end{tabularx}
\end{table}

%% file: paper/05_evaluation.tex
\section{Trustworthy Indic Evaluation}\label{sec:6}

An evaluation is meaningful only if its scores faithfully reflect the capabilities it is intended to measure. For a trustworthy multilingual benchmark, we consider three broad requirements. First, the evaluation tasks themselves must be valid: the questions, expected answers, and scoring procedures should correctly measure the intended capability~\cite{raji2021everything, bowman2021fix, jacobs2021measurement}. Second, when a benchmark is translated from another language, the translation must preserve the underlying task and its evaluation criteria~\cite{artetxe2020artifacts, singh2025globalmmlu, zeng2025marcobench}. Third, multilingual evaluation should, where relevant, reflect the linguistic and cultural context in which the model is expected to operate~\cite{romanou2025include, singh2025globalmmlu, ying2025disentangling}.

Constructing such benchmarks across many languages, cultures, domains, and capabilities, however, requires substantial human effort and is therefore expensive \cite{romanou2025include, singh2025globalmmlu, verma2025milu}. We thus treat \emph{translation as a practical first step rather than an end goal}. Translating established benchmarks allows us to evaluate a broad set of capabilities across languages while retaining common tasks and scoring procedures \cite{lai2023okapi, shi2023mgsm, bandarkar2024belebele, xuan2025mmluprox}. At the same time, we emphasize that translated benchmarks \emph{cannot} replace human-authored evaluations designed specifically for India-centric knowledge, culturally grounded reasoning, or locally relevant tools and environments.

In this section, we describe our attempt to construct an initial slice of a trustworthy multilingual evaluation set. We view this as one practical approach rather than a complete solution, since important human-authored benchmarks covering culturally grounded knowledge, reasoning, and other locally relevant capabilities are still lacking. We first discuss the key challenges and pitfalls involved in multilingual benchmark construction in Section~\ref{sec:6.1}. We then describe our capability taxonomy and benchmark selection in Section~\ref{sec:6.2}. Finally, we introduce a rigorous human-in-the-loop translation and validation pipeline in Section~\ref{sec:6.3}.

\subsection{Challenges in Multilingual Benchmark Construction}
\label{sec:6.1}
At first glance, extending benchmark coverage to multiple languages appears straightforward. One can either translate an existing benchmark while preserving its tasks and evaluation protocol \cite{lai2023okapi, shi2023mgsm, xuan2025mmluprox, he2024multiif}, or replicate its construction methodology in the target language \cite{romanou2025include, verma2025milu, devane2025bhashabench}. In practice, both approaches can compromise evaluation validity if not done carefully. A model's observed score may be affected not only by its underlying capability, but also by errors in the underlying source data, artifacts introduced during data collection or translation, and mismatches between the benchmark context and the setting in which the target language is actually used. Consequently, a multilingual benchmark is meaningful only if these errors are carefully identified and controlled.

\subhead{Errors in Source Instances}
We first manually audit two existing benchmarks, \textbf{MILU} \cite{verma2025milu} and \textbf{IndicIFEval} \cite{jayakumar2026indicifeval}, to understand the errors that may arise even before translation. MILU evaluates broad knowledge and reasoning using questions collected from Indian examinations, whereas IndicIFEval evaluates instruction-following behavior using both natively authored and translated prompts. Together, these benchmarks represent distinct tasks and benchmark construction pipelines.

Our analysis of MILU shows that collecting naturally occurring questions from online sources does not, by itself, guarantee evaluation validity. We find questions with missing figures or passages, incomplete statements, mixed-language content, and incorrect or inconsistent metadata. For example, some questions refer to a chart or figure that is absent from the released instance, while others provide answer choices referring to numbered statements that are missing from the question. Such instances may be syntactically well-formed yet remain impossible to answer from the information provided \cite{gema2025mmlu,vendrow2025reliability}. We summarize representative failure modes in Table~\ref{tab:audit-errors}.

\begin{table}[htbp]
\centering
\caption{Summary of errors found in the MILU and IndicIFEval audits, by type. \emph{Dropped}: rows removed because they could not be fixed. \emph{Repaired}: rows corrected and retained. IndicIFEval errors are in its constraint checkers rather than its questions, so they were all repaired.}
\label{tab:audit-errors}
\small
\begin{tabular}{llrr}
\toprule
\textbf{Benchmark} & \textbf{Error Type} & \textbf{\# dropped rows} & \textbf{\# repaired rows} \\
\midrule
MILU & Incomplete question & 2,234 & 37 \\
MILU & Translation defect & 525 & 482 \\
MILU & Missing figure or passage & 480 & 0 \\
MILU & Malformed question & 172 & 232 \\
MILU & Duplicate options & 156 & 0 \\
IndicIFEval & Silently unenforced constraint & 0 & 2171 \\
IndicIFEval & Unicode word-boundary failure & 0 & 531 \\
IndicIFEval & Non-ASCII letter substitution & 0 & 379 \\
IndicIFEval & Evaluator-prompt mismatch & 0 & 107 \\
\bottomrule
\end{tabular}
\end{table}

\begin{figure}[htbp]
\centering
\includegraphics[width=\linewidth]{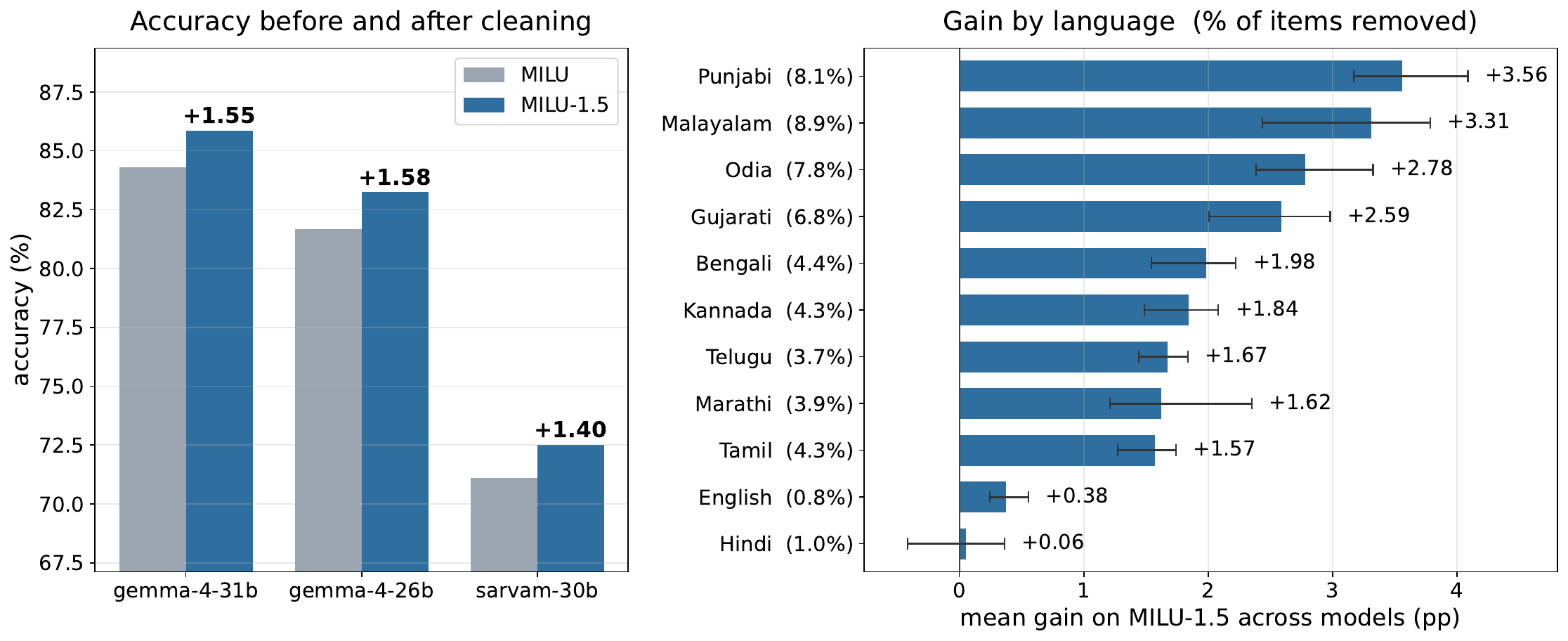}
\caption{Benchmark cleaning shifts accuracy in proportion to how unclean each language set was. Left: 0-shot accuracy on MILU vs. MILU-1.5; all three models gain 1.4–1.6 points. Right: mean gain per language across models, with the share of entries fixed in brackets.}
\label{fig: milu1.5}
\end{figure}

Based on this audit, we release \textbf{MILU-1.5}\footnote{\url{https://huggingface.co/datasets/ai4bharat/MILU-1.5}}, a revised version of MILU in which the identified issues are systematically reviewed and corrected. To quantify the impact of this cleaning, we evaluate gemma-4-31B-it, gemma-4-26B-A4B-it and the sarvam-30b models on both versions and compare the resulting scores, as shown in Figure~\ref{fig: milu1.5}. We observe measurable differences after cleaning, showing that a relatively small number of invalid or ambiguous instances can affect the conclusions drawn from benchmark scores.

Our audit of IndicIFEval reveals a complementary class of problems, summarized in Table~\ref{tab:audit-errors}. At the schema level, the benchmark appears almost entirely clean. However, jointly inspecting the prompts, metadata, and \textbf{released evaluation code} reveals several substantive issues, including evaluator--prompt mismatches, constraints that are silently unenforced, Unicode-sensitive checker failures, and residual artifacts from generation or data collection. Particularly concerning are cases in which English-derived evaluators make assumptions that do not hold for Indic scripts. For example, word-boundary checks fail for some Indic strings, while a letter-frequency checker silently replaces non-ASCII characters with random Latin letters. In such cases, the evaluator still produces a valid-looking score, but effectively checks a different constraint from the one presented to the model.

These audits highlight that \textbf{schema validity is not equivalent to evaluation validity}: a benchmark may be structurally complete while containing unanswerable questions or incorrectly implemented evaluators. Moreover, such issues are not limited to translated data, as the natively authored subset of IndicIFEval contains substantially more flagged instances than its translated subset. We therefore encourage the community to treat \emph{source-benchmark auditing as an important consideration when developing and expanding multilingual benchmarks}.

\subhead{Translation Errors}
A correct source instance may not remain correct after translation. To study such errors, we audit ARC-Challenge \cite{clark2018arc} and MBPP \cite{austin2021mbpp}, two benchmarks with substantially different task structures. As a simple baseline, we translate their instances using frontier LLMs with a generic translation prompt, with minimal benchmark-specific guidance, and inspect cases where translation changes the task being evaluated. Despite generally fluent outputs, we observe several recurring failure modes, which are summarized in Table~\ref{tab:translation-failures}.

For \textbf{ARC-Challenge}, translation can alter linguistic cues and answer distinctions that determine the correct choice. Terms such as \emph{except}, \emph{not}, or \emph{least} may be weakened, while opposing answer choices may be translated too similarly, collapsing distinctions between distractors. \textbf{MBPP} presents a related challenge because natural-language descriptions are tightly coupled with executable semantics. Technical terms such as \emph{substring}, \emph{subsequence}, \emph{list}, and \emph{tuple} must retain precise meanings, while regular expressions, quoted literals, and other code-like spans often need to remain unchanged. Translating these loosely can therefore change the program being specified. We also find cases where frontier LLMs attempt to \textbf{repair inconsistencies in the source instance} rather than faithfully preserve them, for example by rewriting a description to better match the accompanying code. Although such corrections may improve an individual instance, they break equivalence with the source benchmark.

Overall, fluent translation does not guarantee evaluation equivalence. We therefore view benchmark translation as preserving an \textbf{evaluation contract}~\cite{jacobs2021measurement, raji2021everything}: the translated instance must retain what is being asked, what constitutes the correct answer, which distinctions matter for evaluation, and which spans must remain unchanged.

\subhead{Not every capability should be evaluated through translation}
Even a perfect translation does not make every benchmark appropriate for every linguistic setting. Translation changes the language of an instance while largely preserving the world in which it was originally written. This distinction matters because \textbf{cross-lingual comparability and contextual relevance are different evaluation objectives}. Translated benchmarks are useful for testing whether models can solve the same task across languages, but they may not reflect the knowledge, entities, conventions, and situations encountered by speakers of those languages.

\begin{table}[t]
\centering
\footnotesize
\setlength{\tabcolsep}{4pt}
\caption{Failure modes observed when translating ARC-Challenge and MBPP with a generic LLM prompt. Each row gives a source example (with instance ID where available), how direct translation can alter it, and the effect on evaluation.}
\label{tab:translation-failures}
\begin{tabularx}{\linewidth}{@{}
  >{\hsize=0.80\hsize\raggedright\arraybackslash}X
  >{\hsize=1.25\hsize\raggedright\arraybackslash}X
  >{\hsize=1.15\hsize\raggedright\arraybackslash}X
  >{\hsize=0.80\hsize\raggedright\arraybackslash}X@{}}
\toprule
\textbf{Failure type} & \textbf{Source example} & \textbf{Translation risk} & \textbf{Effect on evaluation} \\
\midrule
\rowcolor{gray!12}
\multicolumn{4}{@{}l}{\textbf{ARC-Challenge}} \\
\addlinespace[3pt]
Negation / polarity loss
  & ``Scientists consider all of the following \emph{except}\,\ldots'' (\texttt{Mercury\_184975})
  & \emph{Except} or similar cues (\emph{not}, \emph{least}, \emph{unlikely}) may be weakened or omitted
  & Reverses which option is correct \\
\addlinespace[5pt]
Option distinction collapse
  & \emph{decrease / increase / remain the same} (\texttt{NYSEDREGENTS\_2012\_8\_42})
  & Opposing choices may receive similar or identical translations
  & Distinct distractors become indistinguishable \\
\addlinespace[5pt]
Technical terminology collapse
  & \emph{observation / prediction / conclusion / theory / hypothesis} (\texttt{TIMSS\_2007\_8\_pg53})
  & Closely related scientific terms may be translated with the same everyday word
  & Changes the conceptual distinction being tested \\
\addlinespace[5pt]
Symbol corruption
  & \texttt{R}, \texttt{rr}, \texttt{Rr}, \texttt{RR} (\texttt{Mercury\_7238945})
  & Case-sensitive symbols may be transliterated or normalized in scripts without case
  & Genetically distinct choices collapse \\
\addlinespace[5pt]
Formula / unit modification
  & $\mathrm{2H_2O(l) \rightarrow 2H_2(g) + O_2(g)}$; 32\,\textdegree F
  & Formulae, signs, units or capitalization may be altered or localized
  & Changes the scientific content \\
\addlinespace[5pt]
\midrule
\rowcolor{gray!12}
\multicolumn{4}{@{}l}{\textbf{MBPP}} \\
\addlinespace[3pt]
Programming-term mistranslation
  & \emph{substring}, \emph{subsequence}, \emph{list}, \emph{tuple}, \emph{set}, \emph{dictionary}
  & Related technical terms may be translated as generic equivalents
  & Changes the required algorithm or data structure \\
\addlinespace[5pt]
Literal / regex corruption
  & \texttt{ab\{2,3\}}, \texttt{[A-Z]}, \texttt{'a'}, \texttt{'b'}
  & Quoted characters, regex syntax or ranges may be translated or normalized
  & Changes the program specification \\
\addlinespace[5pt]
Source error silently repaired
  & \emph{dictionary} although the implementation expects a \emph{list}; \emph{all divisors} although the code uses proper divisors
  & A capable LLM may infer the intended task and rewrite the source during translation
  & Translated instance no longer matches the original \\
\bottomrule
\end{tabularx}
\end{table}

Prior work shows that this mismatch can materially affect evaluation. Tomar et al. \cite{tomar2025mathculture}, for example, culturally adapt GSM8K \cite{cobbe2021gsm8k} problems to different regions while preserving their mathematical structure and observe consistent performance differences between the original and localized versions; similar effects have been reported independently \cite{karim2025lostcultural, azime2025culturegap}. Similarly, \textbf{Global MMLU}~\cite{singh2025globalmmlu} finds substantial Western-centric content in MMLU and shows that model rankings can change on culturally sensitive subsets. Ying et al. \cite{ying2025disentangling} further report a \textbf{cultural-linguistic synergy}, where models perform better when the language of a question aligns with its cultural context.

These findings motivate evaluations that begin from the target context rather than only translating English benchmarks. In the Indic setting, benchmarks such as MILU \cite{verma2025milu} source questions from Indian examinations \cite{devane2025bhashabench, patel2025indicparam}, while BharatBBQ \cite{tomar2025bharatbbq} adapts social categories and situations from BBQ \cite{parrish2022bbq} to the Indian sociocultural context. We therefore view \textbf{translated, localized, and natively authored benchmarks as complementary}: translation supports controlled cross-language comparison\cite{xuan2025mmluprox, shi2023mgsm}, localization adapts context while retaining much of the original task structure~\cite{azime2025culturegap, simon2024lofti}, and native construction is necessary when the capability itself depends on local knowledge or conventions~\cite{romanou2025include, maji2025sanskriti, seth2024dosa}. A trustworthy multilingual evaluation should combine these approaches where appropriate, rather than treating translation as a universal solution.

\subsection{Capability taxonomy and benchmark selection}\label{sec:6.2}
Our previous discussion hints towards the fact that multilingual evaluation should begin with the capability being measured. However, constructing a comprehensive capability taxonomy together with natively authored evaluations for every capability is beyond the scope of this release. We therefore take a pragmatic first step: we identify a broad set of capabilities commonly evaluated for contemporary general-purpose LLMs~\cite{liang2023helm, gao2024lmeval} and, for each capability, select a combination of English benchmarks that can be meaningfully translated and benchmarks that are already available in multilingual form.

Table~\ref{tab:taxonomy} summarizes the current taxonomy and benchmark set. We do not intend this list to be exhaustive. Several important capabilities remain absent or only partially represented, particularly those that require locally grounded knowledge \cite{romanou2025include, maji2025sanskriti}, tools, environments, or interaction settings \cite{yao2024taubench, barres2025tau2bench} and therefore cannot be adequately evaluated through translation alone.

\begin{table}[htbp]
\centering
\small
\renewcommand{\arraystretch}{1.6}
\caption{Capabilities and benchmarks in the Indic-LLM-Eval-Suite.}
\label{tab:taxonomy}
\begin{tabularx}{\linewidth}{@{}p{0.35\linewidth}>{\raggedright\arraybackslash}X@{}}
\toprule
\textbf{Capability} & \textbf{Benchmarks} \\
\midrule
General chat          & Arena Hard~\cite{li2024arenahard}, BiGGen Bench~\cite{kim2025biggenbench}, AlpacaEval~\cite{dubois2024lcalpacaeval}, IndiVibe~\cite{sarvam_indivibe} \\
Instruction following & IndicIFEval~\cite{jayakumar2026indicifeval} \\
Math                  & GSM8K~\cite{cobbe2021gsm8k}, MATH-500~\cite{hendrycks2021math}, AIME 2026~\cite{matharena_aime2026}, FrontierMath (sample)~\cite{glazer2024frontiermath} \\
STEM                  & JEEBench~\cite{arora2023jeebench}, GPQA~\cite{rein2024gpqa}, HLE~\cite{phan2025hle} \\
Commonsense reasoning & ARC-Challenge~\cite{clark2018arc} \\
Knowledge             & MMLU-Pro~\cite{wang2024mmlupro}, MILU-1.5~\cite{verma2025milu}, BhashaBench-multi~\cite{devane2025bhashabench}, IndicQuest-v2~\cite{jain2026l3cube} \\
Code                  & HumanEval~\cite{chen2021humaneval}, MBPP~\cite{austin2021mbpp}, SWE-bench~\cite{jimenez2024swebench} \\
Translation           & IN22~\cite{gala2023indictrans2}, WMT24++~\cite{deutsch2025wmt24pp}, Indic Translation Bench \\
\bottomrule
\end{tabularx}
\end{table}

Our selection is intentionally biased toward benchmarks that are already available and whose underlying tasks remain meaningful across languages when translated carefully. Capabilities such as mathematical reasoning and scientific question answering, for example, often permit relatively direct cross-lingual evaluation when their semantics and scoring criteria are preserved. The table should therefore be interpreted as a \textbf{work in progress}, rather than a complete Indic-LLM Evaluation Suite. It reflects what can be evaluated reliably with existing resources today, while leaving substantial room for benchmarks that must be localized more deeply or developed from scratch. The suite should evolve as new benchmarks become available, existing benchmarks are reassessed, and new evaluations are developed for capabilities where translation provides an insufficient or misleading signal.

\subsection{The Translation Pipeline}\label{sec:6.3}
\begin{figure}[htbp]
\centering
\includegraphics[width=\linewidth]{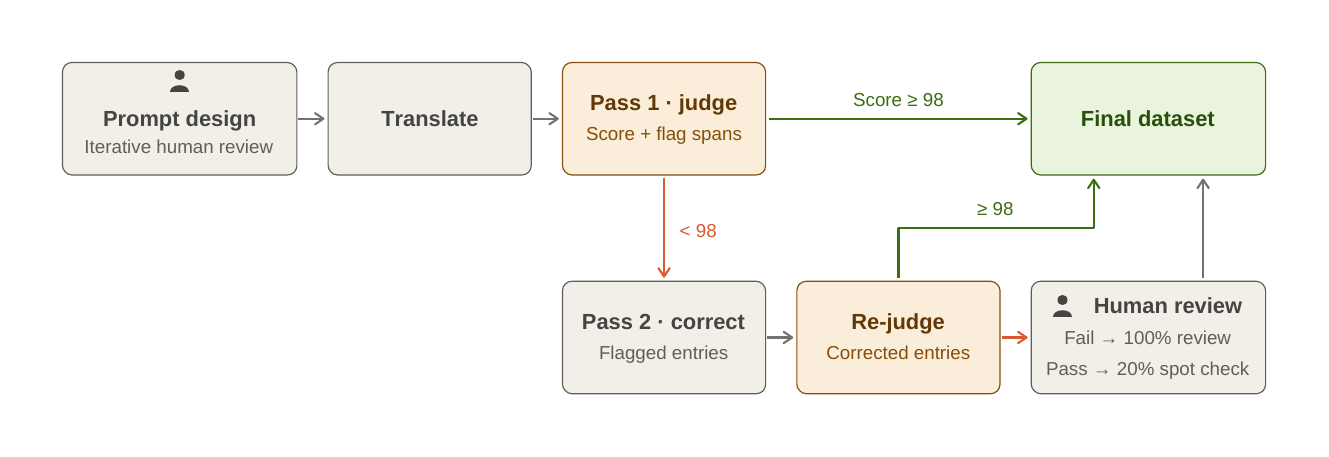}
\caption{Two-pass translation quality control. An LLM judge scores entries (pass ≥ 98) and flags error spans. Failed entries are corrected and re-judged. Remaining failures get 100\% human review, and 20\% of corrected passes are spot-checked.}
\label{fig:translation-pipeline}
\end{figure}

Translating a benchmark while preserving its evaluation properties requires more than generating fluent target-language text. As discussed in Section~\ref{sec:6.1}, different benchmarks contain different task-specific nuances that must be preserved during translation. We therefore construct each translated benchmark using the iterative human-in-the-loop pipeline shown in Figure~\ref{fig:translation-pipeline}~\cite{singh2025globalmmlu, zeng2025marcobench, he2024multiif}. For all stages, we use Frontier LLMs for translation.

Starting from an English benchmark, we first develop benchmark-specific translation, judging, and correction prompts through iterative human review. The finalized translation prompt is then used to generate target-language instances, which undergo structural validation followed by LLM-based evaluation using a benchmark-specific rubric. Instances that do not meet the quality threshold are passed to a targeted correction stage and subsequently re-evaluated using the same procedure. This \textbf{translate → evaluate → correct → evaluate} cycle allows us to systematically identify and repair translation errors while preserving the structure and evaluation protocol of the source benchmark.

\subsubsection{Human-in-the-loop Prompt Development}\label{sec:6.3.1}
A single generic translation prompt cannot capture the task-specific requirements of the diverse benchmarks in our suite. Multiple-choice, mathematical reasoning, and code-generation tasks, for example, contain different components that must either be translated carefully or preserved exactly. We therefore develop separate \textbf{translation, judging, and correction prompts} for each benchmark. For each benchmark, we first analyze its structure, evaluation protocol, and translation-sensitive components to construct an initial prompt. We apply this prompt to a small sample and manually inspect the outputs for task-specific errors. Based on these observations, we iteratively refine the prompt until the major recurring failure modes are addressed. We follow the same procedure when developing the corresponding judging and correction prompts.

Across benchmarks, we find that the most important refinements concern \textbf{task preservation rather than linguistic fluency}. Mathematical benchmarks require explicit instructions to preserve numerals and mathematical notation, while code-generation benchmarks require code, identifiers, and executable structures to remain unchanged. Multiple-choice tasks additionally require preserving the number, order, and correspondence of answer options. These requirements form the basis of the benchmark-specific prompts used throughout the remainder of the pipeline.

\subsubsection{Benchmark-aware Translation}\label{sec:6.3.2}
Once the translation prompt is finalized, we translate the complete benchmark. Each request contains the source instance, the target language, and the corresponding benchmark-specific prompt. The model returns a structured representation of the translated instance, with explicit instructions specifying which fields should be translated and which must remain unchanged. Importantly, we translate the \textbf{entire instance jointly in a single request}, rather than translating individual fields or answer choices independently~\cite{artetxe2020artifacts, singh2025globalmmlu}. This helps preserve relationships between different components of the instance, such as the question and its answer options, and reduces errors such as semantically distinct choices collapsing to similar target-language expressions or terminology becoming inconsistent across fields.

Each translated instance then undergoes lightweight programmatic validation to ensure that it is parseable, contains the expected fields, and satisfies benchmark-specific structural constraints, such as field types, the number of answer choices, and their ordering. Instances that pass these checks proceed to the automatic evaluation stage.

\subsubsection{Automatic Translation Evaluation}\label{sec:6.3.3}
We evaluate every translated instance using a benchmark-specific judging prompt based on an \textbf{MQM-style error annotation rubric}~\cite{lommel2014mqm, freitag2021mqm}. In addition to an overall quality score, the judge produces structured error annotations identifying the relevant source and target spans, error category, severity, and a brief explanation~\cite{kocmi2023gembamqm, fernandes2023automqm, guerreiro2024xcomet, kocmi2024esa}. This provides a more actionable signal than a scalar score alone~\cite{fernandes2023automqm, doddapaneni2024fbi} and allows the subsequent correction stage to directly target the identified errors. We use an overall quality score of 98\% as the acceptance threshold. Instances scoring below this threshold are routed to the correction stage. Across the benchmarks evaluated in this release, \textbf{89.8\%} of translated instances satisfy the acceptance criterion on the first pass, while the remaining instances require correction.

\subsubsection{Targeted Correction}\label{sec:6.3.4}
Instances that fail the initial evaluation are passed to a targeted correction stage. The model receives the source instance, the current translation, and the structured error annotations produced during evaluation. Rather than re-translating the entire instance, the correction prompt instructs the model to modify only the identified problematic spans~\cite{lu2024mqmape}, thereby minimizing unnecessary changes to portions of the translation that have already been validated. Each corrected instance is then passed through the same judging procedure before acceptance. Among all the instances sent for correction, \textbf{54.9\%} improve in overall quality score and \textbf{48.5\%} subsequently satisfy the acceptance criterion. The number of instances containing major or critical errors decreases from \textbf{1721} after the initial translation stage to \textbf{170} after correction, indicating that targeted correction substantially reduces the most consequential translation errors.

\subsection{Human Verification}\label{sec:6.4}
Automated evaluation is used to scale quality control, but not as a substitute for human validation \cite{freitag2021mqm, singh2025globalmmlu, hada2024evaluators}. We therefore conduct native-speaker review on two complementary subsets of all the translated benchmarks. First, we review \textbf{all instances that remain below the acceptance threshold of 98 after translation, correction, and re-judging}, including instances for which the automated pipeline does not produce a reliable final judgement. Second, we randomly sample \textbf{20\% of the instances that pass the automated pipeline}. The former provides complete coverage of the difficult tail, while the latter allows us to estimate how often the automated audit incorrectly accepts an invalid translation.

Reviewers are shown the English source instance together with its final translated version and the corrected version where available, and are asked to assess whether the translation preserves the meaning and evaluation intent of the source. Where an issue is identified, reviewers additionally record its type and provide a corrected translation where appropriate. This enables us to measure both the validity of the released benchmark and the agreement between automated and human assessments \cite{watts2024pariksha}. 

Across the reviewed set, 73.1\% of the automatically accepted instances were judged valid by native speakers, suggesting a false-accept rate of 26.9\% for the automated pipeline. Among instances that continued to fail after automated correction, 54.8\% required human modification, while the remaining instances were found to be acceptable despite being rejected by the automated judge. Human reviewers found that the most common remaining errors involved source terminology left untranslated or transliterated rather than rendered, outright mistranslation and omitted or duplicated content. We additionally observe that the human edit rate varies by more than a factor
of two across languages, from 82.0\% of reviewed Punjabi instances to 35.2\%
for Odia, and that judged validity moves independently of it as Nepali has the
highest validity (95.9\%) despite a 60.3\% edit rate, while Tamil (46.6\%) and
Urdu (49.2\%) are lowest, indicating that the residual errors are not distributed uniformly across the suite.

These results provide an independent check on the automated audit-and-repair pipeline. In particular, performance on the randomly sampled passing set measures whether the high automated acceptance rate corresponds to genuine translation validity, while the failed tail identifies the classes of errors that remain difficult to resolve automatically. The final released benchmark incorporates the human corrections from this stage.

\subsection{Evaluating Contemporary Open Models}\label{sec:6.5}
Having constructed and validated the evaluation suite, we next use it to evaluate the capabilities of contemporary open-weight indic-multilingual models. Our goal is not to establish a leaderboard, but to examine how current models perform across the capability categories introduced in Section~\ref{sec:6.2}, how consistently these capabilities transfer across Indic languages, and where substantial gaps remain.

\subsubsection{Evaluation Setup}\label{sec:6.5.1}
We evaluate a diverse set of open-weight models spanning different model families and parameter scales, including \textbf{models from the Nemotron family, google/gemma-4-31B-it,  gnani/gnani-evon-v3.3-30B-A3, meta-models/Muse-Glimmer-30B and Qwen/Qwen3-30B-A3B}. We include both Indic and general multilingual models. 

\clearpage

\begin{table}[p]
\centering
\scriptsize
\setlength{\tabcolsep}{2pt}
\renewcommand{\arraystretch}{1.00}

\begin{tabularx}{\linewidth}{
  @{}
  >{\raggedright\arraybackslash}p{3.4cm}
  @{\hspace{3pt}}
  >{\raggedright\arraybackslash\itshape}p{2.2cm}|
  *{10}{>{\centering\arraybackslash}X}
  @{}
}
\toprule

\textbf{Task}\rule{0pt}{1.0cm}
& \textbf{\upshape }

& \rotatebox{90}{\bfseries Nemotron-3.5-Super-120B}

& \rotatebox{90}{\bfseries Nemotron-3-Super-120B}

& \rotatebox{90}{\bfseries sarvam-105b-1026}

& \rotatebox{90}{\bfseries Gemma-4-31B-it}

& \rotatebox{90}{\bfseries Gnani-evon-v3.3-30B}

& \rotatebox{90}{\bfseries Muse-Glimmer-30B}

& \rotatebox{90}{\bfseries Nemotron-3.5-Lightning-30B}

& \rotatebox{90}{\bfseries Qwen3-30B}

& \rotatebox{90}{\bfseries Sarvam-30b}

& \rotatebox{90}{\bfseries Gemma-4-26B-A4B-it}
\\
\midrule
\rowcolor[gray]{0.94}
\multicolumn{12}{l}{\textbf{Mathematics}} \\

Indic AIME 2026
&pass@4, acc
&\textbf{95.28}
&90.83
&94.44
&94.44
&83.61
&93.89
&71.39
&79.44
&90.83
&93.33\\

Indic MATH500
&{acc\_norm}
&97.20
&92.62
&96.47
&\textbf{97.77}
&92.33
&95.17
&81.10
&92.20
&95.50
&97.32\\

Indic FrontierMath
&{acc\_norm}
&\textbf{57.64}
&20.14
&22.22
&27.08
&---
&--- 
&---
&--- 
&---
&---\\

\midrule

\rowcolor[gray]{0.94}
\multicolumn{12}{l}{\textbf{STEM}} \\

Indic GPQA Diamond
&{avg[8]acc\_norm}
& \textbf{88.23}
& 76.40
& 72.45
& 84.37
& 64.36
& 82.85
& 62.35
& 58.52
& 59.71
& 79.80\\

Indic JEE Bench
& \multicolumn{11}{l}{} \\
\hspace{1em}Chemistry
&{acc\_norm}
&76.35
&75.45
&66.60
&\textbf{85.02}
&65.98
&15.23
&61.17
&71.79
&55.61
&81.55\\
\hspace{1em}Mathematics
&{acc\_norm}
&88.51
&83.02
&73.99
&\textbf{95.88}
&82.41
&31.01
&74.9
&68.86
&70.90
&92.23\\
\hspace{1em}Physics
&{acc\_norm}
&81.06
&77.79
&66.62
&\textbf{90.94}
&74.61
&21.78
&64.97
&74.37
&63.45
&87.65\\

\midrule

\rowcolor[gray]{0.94}
\multicolumn{12}{l}{\textbf{Knowledge}} \\

MILU 1.5
&{acc\_norm}
&\textbf{90.09}
&82.49
&84.35
&87.63
&80.02
&87.23
&68.83
&78.73
&74.67
&85.61 \\

IndicQuest-v2 (En)
&{judge score}
&\textbf{73.87}
&69.84
&59.26
&65.43
&54.74
&63.96
&52.87
&44.05
&45.81
&62.63
\\

IndicQuest-v2 (In)
&{judge score}
&\textbf{66.40}
&56.44
&51.19
&61.78
&44.98
&61.00
&28.23
&34.15
&36.58
&57.57
\\

\midrule

\rowcolor[gray]{0.94}
\multicolumn{12}{l}{\textbf{Code}} \\

Indic HumanEval
&strict
&97.87
&94.92
&97.21
&\textbf{99.70}
&80.34
&96.49
&85.67
&79.83
&93.34
&99.19\\

Indic MBPP
&strict
&\textbf{88.43}
&85.15
&85.12
&85.88
&79.85
&80.88
&81.90
&79.25
&79.25
&85.32 \\

\midrule

\rowcolor[gray]{0.94}
\multicolumn{12}{l}{\textbf{Translation}} \\

WMT24++
&chrF
&45.07
&31.38
&47.98
&\textbf{50.78}
&43.88
&47.21
&23.80
&40.05
&16.35
&49.15 \\

\addlinespace[2pt]

IN22 Gen
& \multicolumn{11}{l}{} \\

\hspace{1em}EN$\rightarrow$IN
& chrF++
&42.99
&34.52
&47.05
&\textbf{49.67}
&42.4
&46.82
&24.07
&36.92
&17.15
&48.68 \\

\hspace{1em}IN$\rightarrow$EN
& chrF++
&63.41
&56.94
&59.22
&\textbf{64.56}
&59.26
&61.77
&45.67
&58.03
&34.41
&63.03\\

IN22 Conv
& \multicolumn{11}{l}{} \\

\hspace{1em}EN$\rightarrow$IN
& chrF++
&40.38
&32.07
&45.85
&47.44
&37.75
&42.06
&21.86
&34.02
&06.53
&\textbf{47.66} \\

\hspace{1em}IN$\rightarrow$EN
& chrF++
&56.4
&49.73
&54.83
&\textbf{58.72}
&53.14
&55.67
&40.81
&53.13
&12.23
&57.67\\

Indic Translation Bench
& \multicolumn{11}{l}{} \\

\hspace{1em}EN$\rightarrow$IN
&judge score
&33.64
&7.48
&\textbf{81.50}
&58.69
&17.01
&59.81
&3.55
&10.82
&2.80
&52.52\\

\hspace{1em}IN$\rightarrow$EN
&judge score
&51.69
&12.55
&\textbf{68.54}
&64.42
&28.46
&74.34
&4.68
&13.86
&1.12
&46.07\\

\addlinespace[2pt]

\midrule

\rowcolor[gray]{0.94}
\multicolumn{12}{l}{\textbf{IF and Chat}} \\

Indic IFeval 1.5
& \multicolumn{11}{l}{} \\

\hspace{1em}Prompt strict
&
&61.14
&63.18
&82.90
&84.96
&50.34
&\textbf{85.34}
&53.04
&66.66
&45.45
&83.32 \\

\hspace{1em}Instruction strict
&
&70.74
&71.16
&86.75
&89.68
&62.27
&\textbf{89.78}
&61.86
&75.82
&56.47
&88.39 \\

\addlinespace[2pt]

Indic Arena-Hard
& \multicolumn{11}{l}{} \\

\hspace{1em}Correctness
& judge score
&3.12
&2.57
&3.13
&\textbf{3.77}
&2.42
&3.62
&2.44
&2.32
&1.94
&3.67 \\

\hspace{1em}Relevance
& judge score
&3.67
&3.24
&3.80
&\textbf{3.98}
&3.26
&3.95
&3.08
&3.12
&2.91
&3.96 \\

\hspace{1em}Completeness
& judge score
&3.50
&3.02
&3.64
&\textbf{3.89}
&2.74
&3.82
&2.82
&2.76
&2.59
&3.82 \\

\hspace{1em}Language quality
& judge score
&2.43
&1.40
&3.34
&\textbf{3.96}
&1.87
&3.84
&1.77
&2.49
&2.51
&3.93 \\

Indic Alpaca-Eval
& \multicolumn{11}{l}{} \\

\hspace{1em}Instruction following
& judge score
&3.81
&2.74
&4.38
&\textbf{4.91}
&2.52
&4.87
&2.62
&3.93
&3.56
&4.89 \\

\hspace{1em}Correctness
& judge score
&3.76
&3.08
&4.05
&\textbf{4.89}
&4.33
&4.74
&3.09
&3.49
&2.86
&4.85 \\

\hspace{1em}Usefulness
& judge score
&3.46
&2.29
&4.12
&\textbf{4.90}
&3.02
&4.82
&2.48
&3.53
&3.01
&4.88 \\

\hspace{1em}Language quality
& judge score
&2.79
&1.66
&3.64
&\textbf{4.92}
&1.65
&4.84
&1.61
&3.45
&2.88
&\textbf{4.92} \\

IndiVibe
& \multicolumn{11}{l}{} \\
\hspace{1em}Usefulness
& judge score
&52.2
&28.5
&\textbf{83.0}
&77.6
&22.9
&65.3
&27.2
&31.4
&63.1
&72.1 \\

\hspace{1em}Fluency
& judge score
&49.5
&23.9
&80.1
&\textbf{80.3}
&26.2
&65.4
&24.2
&31.9
&60.6
&74.4 \\

\hspace{1em}Language + script
& judge score
&48.0
&25.7
&\textbf{79.7}
&75.8
&25
&62.8
&21.6
&30.0
&64.7
&71.1 \\

\hspace{1em}Verbosity
& judge score
&52.0
&31.0
&\textbf{83.1}
&75.5
&20.4
&63.0
&28.1
&32.0
&67.5
&71.0 \\

\bottomrule
\end{tabularx}
\par\vspace{3pt}
\vspace{4pt}
\caption{Zero-shot comparison of post-trained models across
mathematics, STEM, knowledge, code, translation, and
instruction-following benchmarks. Best available results are marked in \textbf{bold}. }
\label{tab:all-post-trained-results}
\end{table}

\clearpage

All benchmarks in our suite are integrated into \textbf{NeMo-Gym}, which we use as the common evaluation harness. Each benchmark is implemented with a standardized interface covering prompt construction, model invocation, answer extraction, and scoring. This allows the same model configuration to be evaluated consistently across languages and benchmark families, while retaining benchmark-specific scoring procedures. We release these benchmark implementations as part of the evaluation suite to simplify reproduction and future model evaluation.

For benchmarks with deterministic answers, we report the task’s native metric, such as accuracy or pass rate. For open-ended benchmarks, we use the corresponding benchmark-specific evaluation protocol. To make results comparable across capabilities, we additionally report normalized capability-level aggregates computed over the constituent benchmarks. In the Table~6.4 \textit{pass}@\textit{4} measures success within four attempts, \textit{acc\_norm} denotes format-normalized accuracy required for indic languages, \textit{avg[8]} indicates averaging over eight runs, and \textit{strict} code scoring applies no format normalization. Indic Arena Hard, Indic Alpaca Eval and IndiVibe evaluations use Gemini 3.8 Flash for pointwise rubric scoring, but their scales are not comparable. Additionally WMT++ scores are reported only on Indic subset.

\subsubsection{Overall Capability Profile}\label{sec:6.5.2}
Table~6.4 summarizes the performance of the evaluated models across the capability taxonomy. Several broad trends emerge. First, performance varies substantially across capabilities. The strongest models perform well on standard mathematical reasoning, knowledge, and code-generation tasks, with scores above 95 on Indic MATH500 and near 100 on Indic HumanEval. Advanced mathematical reasoning remains considerably more difficult, as illustrated by the 12-question public Indic FrontierMath sample. Translation and open-ended interaction also remain challenging.

Second, scaling generally improves performance, but the magnitude of improvement differs considerably across capabilities. For example, sarvam-105b-1026 improves over Sarvam-30B from 59.71\% to 72.45\% on normalized Indic GPQA and from 16.35 to 47.98 chrF on WMT24++, whereas their Indic MATH500 scores differ by less than one percentage point. Similar variation appears within the Nemotron family, indicating that additional scale and post-training do not yield uniform gains across tasks.
Using a capability-balanced aggregation of the metrics reported in Table~6.4, Gemma-4-31B-IT, Nemotron-3.5-Super-120B and sarvam-105b-1026 achieve the strongest overall performance. This aggregate nevertheless hides substantial variation. The Nemotron-3.5-Super-120B achieves the strongest capability-level averages in mathematics, STEM, knowledge, and code, but trails the Gemma models substantially in translation, instruction following, and open-ended chat. We therefore avoid treating the aggregate score as a complete measure of multilingual capability and instead analyze performance along the individual dimensions below.

\subsubsection{Performance Across Languages}\label{sec:6.5.3}
A central question is whether improvements observed in English transfer uniformly to Indic languages. Among our evaluations, MILU 1.5 provides the directly comparable English and Indic-language scores needed for this analysis. English performance exceeds the average Indic-language performance for all 10 models with completed MILU results. The strongest models reduce this gap considerably.
Importantly, cross-lingual variation is not uniform across capabilities. On Indic MATH500, the strongest models vary by no more than approximately three percentage points across the twelve evaluated Indic languages. Instruction following is less consistent: the per-language strict-prompt scores of the leading Gemma, Muse, and Nemotron-3.5-Super-120B models span approximately 8.5–11.6 percentage points. Multilingual performance therefore cannot be characterized adequately by a single language-level average; the interaction between language and capability is itself an important dimension of evaluation.
Indic-specialized models are competitive on selected tasks but do not consistently outperform broadly multilingual models. Indic specialization can provide strong performance on particular tasks, but linguistic specialization alone does not guarantee stronger reasoning, instruction-following, or interaction capability across the suite.

\subsubsection{Summary}\label{sec:6.5.4}
Overall, our evaluation reveals substantial variation across both capabilities and languages. While the
strongest open models achieve competitive performance on several translated benchmarks, significant
gaps remain on advanced reasoning, agentic tasks, safety and Indic local knowledge. Moreover,
performance on translated tasks does not always predict performance on Indic-specific evaluations,
emphasizing the need for both forms of measurement.
These results highlight two broader conclusions. First, multilingual model quality should be
evaluated across multiple capabilities and languages rather than summarized by a small set of
academic benchmarks. Second, the remaining performance gaps motivate continued development
of both stronger multilingual models and evaluation tasks that more closely reflect the contexts in
which Indic-language systems will ultimately be used.

%% file: paper/12_conclusion.tex
\section{Conclusion}\label{sec:12}

Indic languages remain far behind English in both pretraining data and trustworthy evaluation. We began this work from two principles: that foundational resources, benchmarks and recipes should be open and reproducible, and that LLM-driven translation, paired with automated multi-stage quality checks and targeted human verification, can act as a force-multiplier for extending English-language knowledge to many languages at once. This report applies both principles across all the 22 Indic languages covered in this work.

We make four core contributions. First, we release a source-traceable synthetic pretraining corpus covering all 22 languages in native, with romanized subsets for twelve languages. Second, we release the reusable pipeline that produced it, covering translation, quality assessment, targeted repair and romanization, together with a record of its failure modes. Third, controlled continued-pretraining experiments show that this data improves Indic performance beyond what the best existing open corpora achieve, without measurably harming English. Fourth, we release the Indic-LLM-Eval-Suite, which combines audited source benchmarks with benchmark-aware translation and native-speaker validation, and is integrated into NeMo-Gym for standardized evaluation.

Taken together, these results support the principles we started from, with qualifications. Translation is a strong force-multiplier for general knowledge and reasoning, but it cannot supply facts that the English source never contained. Its reliability comes from checks that are themselves measured against people. Openness is what makes those checks auditable, and several of our findings came from the community's own released benchmarks and evaluators.

Looking ahead, we will continue to improve the Indic-LLM-Eval-Suite, adding more benchmarks and capabilities as existing evaluations saturate and new needs emerge. We will also explore other ways to expand pre-training corpora for Indian languages, beyond translation of English sources. Promising directions include rephrasing and restructuring natively authored Indic text directly in the target language, persona-based data generation, generation grounded in Indian curricula and source documents to supply the India-specific knowledge that translation cannot, and natively generated reasoning data. We plan to build on this corpus in future Nemotron models, so that Indic language support becomes a core part of how these models are trained rather than an afterthought.

We close with a call to the community. Translated resources can only be a first step: the field also needs natively authored, culturally grounded data and benchmarks, especially for the lower-resource languages, and native speakers are the scarcest and most valuable input to that work. We ask researchers to audit the benchmarks and report defects, to contribute new tasks and languages to the Indic-LLM-Eval-Suite through NeMo-Gym. No single organization can cover the linguistic and cultural breadth of India. By releasing the corpus, recipes and evaluation tooling openly, we hope to make sure that effort is shared and cumulative rather than repeated.

%% file: paper/appendix_A_prompts.tex
\appendix
\setcounter{section}{0}
\section{Prompts of the Indic SDG pipeline}\label{sec:A}

The Indic SDG pipeline of Section~\ref{sec:bharat-sdg-pipeline} is driven by a small set of prompts: whole-document translation (plain-text and JSON-output variants), the quality-assessment judge, targeted rewriting of low-scoring translations, native-numeral variants of the translation and judge prompts, and the romanization prompt with its script-aware judge. We do not reproduce them here. All prompts, together with the per-language configurations, are released with the pipeline in \url{https://github.com/NVIDIA-NeMo/DataDesignerPlugins/tree/main/plugins/data-designer-nemotron-indic-translation}, so that every configuration in Table~\ref{tab:3.1} can be regenerated with the exact instructions used in production.

%% file: paper/appendix_corpus.tex
\section{Corpus Size and Accounting}
\label{app:corpus-accounting}

Table~\ref{tab:3.1} reports several quantities for describing the size and yield of the released corpus. Because the corpus is constructed by translating English source documents into Indic languages, no single token count fully characterizes its scale. We therefore distinguish between source tokens, target tokens, document counts, and the expansion ratio.

\paragraph{Source tokens.}
We define \emph{source tokens} as the number of English tokens in the Nemotron-CC v2 documents corresponding to the released translations. Source tokens are counted with the same Gemma-4 tokenizer as the target tokens, so that the two counts and the expansion ratio below are comparable. Across the complete release, the released documents correspond to approximately 1,640B English source tokens, of which 1,357.6B correspond to the twelve native-script configurations of the dual-script languages (the native-script subtotal of Table~\ref{tab:3.1}).

For the native-script generation pipeline of the dual-script languages, approximately 1,426B source tokens enter the pipeline. After quality assessment and rewriting, 1,357.6B of these source tokens, approximately 95\%, are represented in the final released corpus; the remainder belongs to documents whose translations were rejected. Table~\ref{tab:3.1} reports, for each configuration, the share of these source tokens accepted on the first pass and the target tokens recovered through rewriting.

\paragraph{Target tokens.}
We define \emph{target tokens} as the number of tokens in the released Indic text when tokenized using the Gemma-4 tokenizer. We use this tokenizer consistently for corpus-level accounting because Gemma-4 is used in our primary generation pipeline.

The release contains approximately 2,601B target tokens in native script across the 22 languages. Including the romanized configurations increases the total to approximately 2,826B tokens. Among the dual-script languages, 438B target tokens correspond to documents recovered through the rewriting stage.

We emphasize that token counts depend on the tokenizer used and should therefore not be interpreted as tokenizer-independent measures of corpus size. Counts obtained using other tokenizers may differ substantially, particularly for scripts that receive limited coverage in the tokenizer vocabulary.

\paragraph{Expansion ratio.}
We define the \emph{expansion ratio} as the number of target tokens divided by the number of English source tokens for the same set of documents. This quantity captures both differences between the source and target languages and the behavior of the tokenizer on the corresponding script.

The expansion ratio therefore should not be interpreted directly as a linguistic measure of translation length. Among the dual-script languages, it ranges from approximately 1.2 for Bengali to 2.9 for Odia. It reaches approximately 7.6 for Manipuri, largely because the Gemma-4 tokenizer segments Meitei Mayek into many small units.

\paragraph{First-pass and rewritten data.}
For configurations using the rewrite pipeline, Table~\ref{tab:3.1} reports two kinds of first-pass figures. Under \emph{Source}, the \emph{1st pass} column gives the percentage of the English source tokens that entered the pipeline whose translations were accepted by the quality judge on the first pass; the remaining source tokens were either recovered through rewriting or rejected and excluded from the release. Under \emph{Target}, the \emph{1st pass} and \emph{Rewrite} columns give the target tokens of the documents accepted on the first pass and of those recovered through the rewriting stage; their sum is the \emph{Total} column. Languages processed without a rewrite stage report the total only.

\paragraph{Document counts.}
Finally, \emph{Docs} denotes the number of documents in the final released configuration after quality filtering, rewriting where applicable, and final deduplication. Across all 34 configurations, the release contains approximately 2.44B documents. We describe the final filtering and deduplication procedure in Appendix~\ref{app:bharat-sdg-execution}.

For romanized configurations, the documents are transliterations of subsets of the corresponding accepted native-script documents. Consequently, the native-script and romanized document and token counts should not be summed when estimating the amount of semantically distinct training content in the release.

%% file: paper/appendix_C_translator_selection.tex
\section{Per-language tables: translator selection}\label{sec:C}
\begin{table}[htbp]
\centering
\caption{Sentence-level translation quality of candidate translators, English into 19 Indic languages. IndicGenBench (Flores subset), chrF++ (0 to 100, higher is better). Dogri, Kashmiri and Sindhi are not in the benchmark. Manipuri (mni) is not comparable across systems: the reference is in Bengali script and the dedicated translators emit Meitei Mayek, so it is excluded from the low-resource mean. "Priority 12" = the twelve Gemma-pipeline languages; "Low-res 6" = asm, brx, kok, mai, san, sat. T = sampling temperature. Gemma-4-31B-it rows at T=1 use top-p 0.95. The production configuration for the twelve priority languages is plain text, reasoning off, temperature 1.0, top-p 0.95.}\label{tab:C.1}
\begin{adjustbox}{max width=\linewidth}
\setlength{\tabcolsep}{2.5pt}
\scriptsize
\begin{tabular}{>{\raggedright\arraybackslash}p{110pt}rrrrrrrrrrrrrrrrrrrrrr}
\toprule
\textbf{System} & \textbf{asm} & \textbf{ben} & \textbf{brx} & \textbf{guj} & \textbf{hin} & \textbf{kan} & \textbf{kok} & \textbf{mai} & \textbf{mal} & \textbf{mar} & \textbf{mni} & \textbf{nep} & \textbf{ori} & \textbf{pan} & \textbf{san} & \textbf{sat} & \textbf{tam} & \textbf{tel} & \textbf{urd} & \makecell[l]{\textbf{Mean} \\ \textbf{19}} & \makecell[l]{\textbf{Priority} \\ \textbf{12}} & \makecell[l]{\textbf{Low-res} \\ \textbf{6}} \\
\midrule
IndicTrans2 en-indic 1B & 39.4 & 51.6 & 44.0 & 55.4 & 58.1 & 54.6 & 41.7 & 49.4 & 55.5 & 50.4 & 1.0 & 55.8 & 47.5 & 51.0 & 30.0 & 27.8 & 56.5 & 58.3 & 50.6 & 46.2 & 53.8 & 38.7 \\
Sarvam-translate 4B (T=0) & 40.1 & 49.8 & 45.0 & 55.2 & 57.3 & 52.9 & 41.0 & 45.5 & 53.8 & 49.8 & 0.7 & 53.5 & 50.4 & 51.6 & 29.5 & 32.0 & 53.8 & 56.2 & 50.7 & 45.7 & 52.9 & 38.8 \\
Gemma-4-31B-it (T=1, reasoning off) & 36.6 & 47.0 & 19.5 & 50.6 & 55.5 & 50.1 & 30.3 & 43.1 & 49.8 & 47.7 & 20.0 & 50.4 & 44.8 & 48.7 & 28.2 & 21.6 & 51.0 & 50.6 & 48.5 & 41.8 & 49.6 & 29.9 \\
Gemma-4-31B-it (T=1, reasoning on) & 37.7 & 47.0 & 17.6 & 50.9 & 55.8 & 50.6 & 29.6 & 44.3 & 51.3 & 48.5 & 20.5 & 50.8 & 46.0 & 49.3 & 28.8 & 15.4 & 51.9 & 51.5 & 48.9 & 41.9 & 50.2 & 28.9 \\
Gemma-4-31B-it + spec-decode 4 (T=1, off) & 36.5 & 46.9 & 18.5 & 50.5 & 55.6 & 50.2 & 30.2 & 43.1 & 49.7 & 47.5 & 18.5 & 50.1 & 44.7 & 48.8 & 28.1 & 20.5 & 50.9 & 51.1 & 48.6 & 41.6 & 49.5 & 29.5 \\
Gemma-4-31B-it (T=0, JSON output) & 37.0 & 47.2 & 17.9 & 51.3 & 55.9 & 50.8 & 31.1 & 43.8 & 50.2 & 48.1 & 19.0 & 50.6 & 45.1 & 48.9 & 28.5 & 23.0 & 51.6 & 51.5 & 48.8 & 42.1 & 50.0 & 30.2 \\
Gemma-4-31B-it (T=0, plain text) & 34.8 & 48.6 & 1.0 & 51.7 & 56.9 & 51.8 & 29.5 & 43.1 & 52.0 & 49.4 & 1.4 & 52.3 & 43.8 & 45.2 & 25.0 & 7.8 & 53.2 & 53.4 & 49.7 & 39.5 & 50.7 & 23.5 \\
Gemma-4-31B base (T=1) & 28.3 & 38.7 & 17.2 & 39.9 & 45.3 & 40.5 & 24.3 & 38.5 & 41.3 & 37.9 & 12.1 & 42.1 & 34.2 & 39.1 & 21.7 & 23.0 & 41.5 & 43.6 & 38.4 & 34.1 & 40.2 & 25.5 \\
Gemma-4-26B-A4B base (T=1) & 26.6 & 37.7 & 17.9 & 37.9 & 44.0 & 36.1 & 21.2 & 35.1 & 37.8 & 36.6 & 14.1 & 42.8 & 29.5 & 36.3 & 16.7 & 21.6 & 39.0 & 39.8 & 37.6 & 32.0 & 37.9 & 23.2 \\
Gemma-4-12B-it (T=0, plain text) & 23.9 & 48.1 & n/a & 51.0 & 56.5 & 50.6 & 0.2 & 14.6 & 51.1 & 48.4 & n/a & 51.8 & 31.6 & 41.5 & 13.5 & 0.1 & 52.6 & 52.7 & 48.7 & 37.5 & 48.7 & 10.5 \\
TranslateGemma-27B-it & 33.1 & 42.6 & 1.7 & 47.4 & 51.7 & 46.9 & 29.4 & 40.6 & 45.8 & 43.8 & 0.7 & 46.5 & 38.3 & 45.7 & 25.5 & 16.4 & 49.6 & 48.7 & 45.1 & 36.8 & 46.0 & 24.5 \\
TranslateGemma-12B-it & 30.0 & 41.6 & 0.3 & 46.6 & 50.5 & 43.9 & 26.4 & 36.6 & 41.6 & 42.7 & 0.4 & 42.3 & 33.6 & 42.6 & 26.0 & 4.1 & 48.7 & 47.2 & 42.6 & 34.1 & 43.7 & 20.6 \\
TranslateGemma-4B-it & 18.4 & 41.5 & 10.2 & 45.1 & 49.5 & 40.5 & 23.6 & 32.1 & 37.1 & 42.6 & 2.2 & 39.1 & 15.8 & 1.3 & 19.9 & 0.8 & 48.3 & 47.2 & 37.6 & 29.1 & 37.1 & 17.5 \\
Qwen3.6-27B (FP8) & 30.0 & 45.2 & 0.9 & 41.4 & 52.1 & 47.7 & 29.2 & 32.0 & 41.6 & 44.1 & 8.4 & 43.8 & 34.6 & 37.8 & 18.9 & 0.7 & 46.3 & 49.1 & 46.3 & 34.2 & 44.2 & 18.6 \\
Qwen3.5-35B-A3B (FP8) & 19.8 & 43.9 & 0.7 & 45.6 & 44.9 & 38.5 & 19.2 & 25.3 & 45.4 & 43.5 & 1.2 & 36.6 & 34.9 & 38.3 & 13.9 & 0.5 & 40.0 & 46.8 & 45.3 & 30.8 & 42.0 & 13.2 \\
Nemotron 3 Super 120B-A12B base (T=0) & 12.4 & 34.8 & 1.9 & 29.5 & 54.6 & 26.9 & 12.5 & 33.1 & 24.8 & 30.5 & 1.6 & 35.9 & 24.8 & 30.4 & 15.0 & 2.5 & 31.2 & 24.2 & 36.4 & 24.4 & 32.0 & 12.9 \\
Nemotron 3 Nano 30B-A3B base (T=0) & 1.8 & 15.3 & 1.6 & 6.9 & 47.9 & 6.4 & 4.3 & 9.0 & 6.3 & 11.0 & 1.0 & 12.9 & 4.1 & 8.3 & 4.9 & 2.8 & 6.8 & 8.1 & 13.8 & 9.1 & 12.3 & 4.1 \\
\bottomrule
\end{tabular}
\end{adjustbox}
\par\smallskip\footnotesize Notes. (1) The four IndicTrans2 serving variants (transformers, rotary, ctranslate2 fp16, ctranslate2 adalat) agree to within 0.1 on the mean; the transformers row is shown. (2) The Gemma-4-31B-it plain-text row at T=0 collapses on brx, mni and sat (repetition); the JSON row at the same temperature does not. (3) Reverse-direction (Indic to English) results are not used by the pipeline; there, Gemma-4-31B base at T=0 scores 61.4 and Gemma-4-31B-it 57.2.
\end{table}

%% file: paper/appendix_execution.tex
\section{Execution, efficiency, and corpus assembly}
\label{app:bharat-sdg-execution}
\label{sec:bharat-sdg-assembly}

We schedule generation in batches with NeMo Data Designer and serve the models with vLLM. Translation and quality checks can run concurrently. Speculative decoding, FP8 inference, and attention-kernel optimizations improve processing speed. We compare translation quality before and after inference changes alongside the performance measurements.

In a controlled translation benchmark for the top-12 Indic languages, we reduced median cost from 0.835 to 0.335 GPU-hours per million English input tokens, or about 60\%. Both configurations used the same benchmark, with approximately 46.8 million input tokens per language. This corresponds to 2.49 times as many input tokens per GPU-hour. It measures translation, not the end-to-end cost of judging, rewriting, romanization, or the workflows for the remaining ten languages.

Generation returns candidates and their quality scores. Corpus assembly then applies each path's acceptance policy, removes duplicate records introduced by retries or rewrite processing, and retains the source and quality metadata. This source-ID check removes processing duplicates; it is not semantic deduplication of similar content.

Retained source identifiers and quality metadata let users inspect how individual records were produced and select data for downstream training.

%% file: paper/appendix_analysis.tex
\section{Analysis and Failure Modes}\label{app:analysis}\label{sec:analysis}

Development analyses identified failures in numeral preservation and script consistency that were not fully characterized by aggregate translation scores. We describe these observations below, distinguishing results from development tests and generated candidates from the quality of the final released corpus.

\subsection{Numeral preservation and automated assessment}\label{sec:analysis-numerals}

The numeral policy in Section~\ref{sec:bharat-sdg-translation} distinguishes ordinary prose from technical content: configurations may request native numerals in prose while requiring ASCII digits in code, equations, and URLs. Despite these instructions, the translator sometimes introduced native digits into protected technical spans. A closing instruction emphasizing native Hindi digits contributed to this behavior.

In a prompt comparison, removing that instruction reduced the number of documents flagged for digit corruption from 95 to 79 and the total detected violations from 1,012 to 607. Native-digit use in prose remained at the same reported rate. Mean MQM changed only from 96.0 to 96.1. In this comparison, the aggregate score therefore did not reflect the substantial reduction in detected digit violations.

The evaluator also had difficulty applying the numeral policy. A prompt emphasizing script compliance flagged legitimate mathematical content, while reducing that emphasis caused genuine issues to be missed. These observations show why numeral preservation needs assessment at the level of individual technical spans: an aggregate translation score can obscure specific errors, while an overly restrictive script rule can reject acceptable text.

\subsection{Script contamination across languages and sources}\label{sec:analysis-script}

An early analysis of translations from the \texttt{cc\_v2\_diverse\_qa} source subset found substantial variation in cross-script flags across languages. Static checks flagged 41\% of Odia documents, 30\% of Kannada documents, and 19\% of Punjabi documents, compared with 0.4\% for Hindi, 1.5\% for Marathi, and 3.1\% for Urdu. These figures describe detector flags in the development sample. They do not establish the proportion of translations with substantive errors, since text in another script can include material that should be preserved from the source.

A subsequent review of low-quality candidates identified the \texttt{cc\_v2\_english\_cc} subset as a prominent source of problems. Kannada outputs included Korean words and other unintended foreign-script text. This observation adds a source dimension to the language-level differences: examining only a pooled score for a language can conceal problems concentrated in particular source subsets.

The rewrite stage described in Section~\ref{sec:bharat-sdg-quality} provides a mechanism for recovering low-scoring candidates. However, the available analyses do not quantify how frequently these specific script errors remained after rewriting and final acceptance. The observed candidate failures therefore motivate checks by both language and source subset, but should not be interpreted as residual error rates in the released corpus.

%% file: paper/appendix_contributors.tex
\section{Contributors}\label{app:contributors}

This report is a joint effort of NVIDIA and AI4Bharat. Contributors are listed alphabetically by first name within each organization.

\paragraph{NVIDIA.}
Abhinav Khattar,
Aishwarya Chhabra,
Ameya Velingker,
Anusha Kamath,
Brandon Norick,
Kanishk Singla,
Kiran Praveen,
Markus Kliegl,
Mohammad Qazim Bhat,
Mostofa Patwary,
Niket Agarwal,
Rajpreet Thethy,
Ramani Akkati,
Raviraj Joshi,
Sai Nishanth Padala,
Shuoyang Ding,
Sriharsha Niverty,
Utkarsh Vaidya,
Vitaly Lavrukhin,
Yev Meyer.

\paragraph{AI4Bharat.}
Anoop Kunchukuttan,
Mitesh Khapra,
Mohammed Safi Ur Rahman Khan,
Rudra Murthy,
Sumanth Doddapaneni,
Thanmay Jayakumar,
Ujjwal Sharma.

\section{Acknowledgements}\label{app:acknowledgements}

We thank SuperAnnotate for supporting the validation of the evaluation data and the human-annotation studies, and the AI4Bharat team for their support with annotation and linguistic expertise.